\pdfoutput=1
\documentclass[11pt]{article}

\usepackage[final]{acl}

\usepackage{times}
\usepackage{latexsym}

\usepackage[T1]{fontenc}
\usepackage[utf8]{inputenc}

\usepackage{microtype}

\usepackage{inconsolata}

\usepackage{multirow}
\usepackage{booktabs}
\usepackage[normalem]{ulem}
\useunder{\uline}{\ul}{}
\usepackage{algorithmicx,algorithm}
\usepackage{graphicx}
\usepackage{color}
\usepackage{colortbl}
\usepackage{amsmath}
\usepackage{subcaption}

\title{CMQCIC-Bench: A Chinese Benchmark for Evaluating Large Language Models in Medical Quality Control Indicator Calculation}

\author{Guangya Yu, Yanhao Li, Zongying Jiang, Yuxiong Jin, Li Dai, Yupian Lin, \\ \textbf{Ruihui Hou,
 Weiyan Zhang,
Yongqi Fan,
 Qi Ye\thanks{~~Corresponding authors.},
 Jingping Liu,
 Tong Ruan\footnotemark[1]}
\\
School of Information Science and Engineering, \\ East China University of Science and Technology, Shanghai, China
\\
\texttt{guangyayu@mail.ecust.edu.cn},
\texttt{\{yeh\_qi1125,ruantong\}@ecust.edu.cn} 
}

\begin{document}
\maketitle
\begin{abstract}
Medical quality control indicators are essential to assess the qualifications of healthcare institutions for medical services. With the impressive performance of large language models (LLMs) like GPT-4 in the medical field, leveraging these technologies for the \textbf{M}edical \textbf{Q}uality \textbf{C}ontrol \textbf{I}ndicator \textbf{C}alculation (MQCIC) presents a promising approach. In this work, (1) we introduce a real-world task MQCIC and propose an open-source Chinese electronic medical records (EMRs)-based dataset (CMQCIC-Bench) comprising 785 instances and 76 indicators. (2) We propose a semi-automatic method to enhance the rule representation. Then we propose the Clinical Facts-based Inferential Rule (CF-IR) method that disentangles the clinical fact verification and inferential rule reasoning actions. (3) We conduct comprehensive experiments on 20 representative LLMs, covering general and medical models. Our findings reveal that CF-IR outperforms Chain-of-Thought methods in MQCIC tasks. (4) We conduct an error analysis and investigate the capabilities of clinical fact verification and inferential rule reasoning, providing insights to improve performance in the MQCIC further. The dataset and code is available in this repository~\footnote{\href{https://github.com/YuY-2001/C-MQCIC}{https://github.com/YuY-2001/C-MQCIC}}.

\end{abstract}

\section{Introduction}
\begin{figure}[t]
\centering
\includegraphics[width=\linewidth]{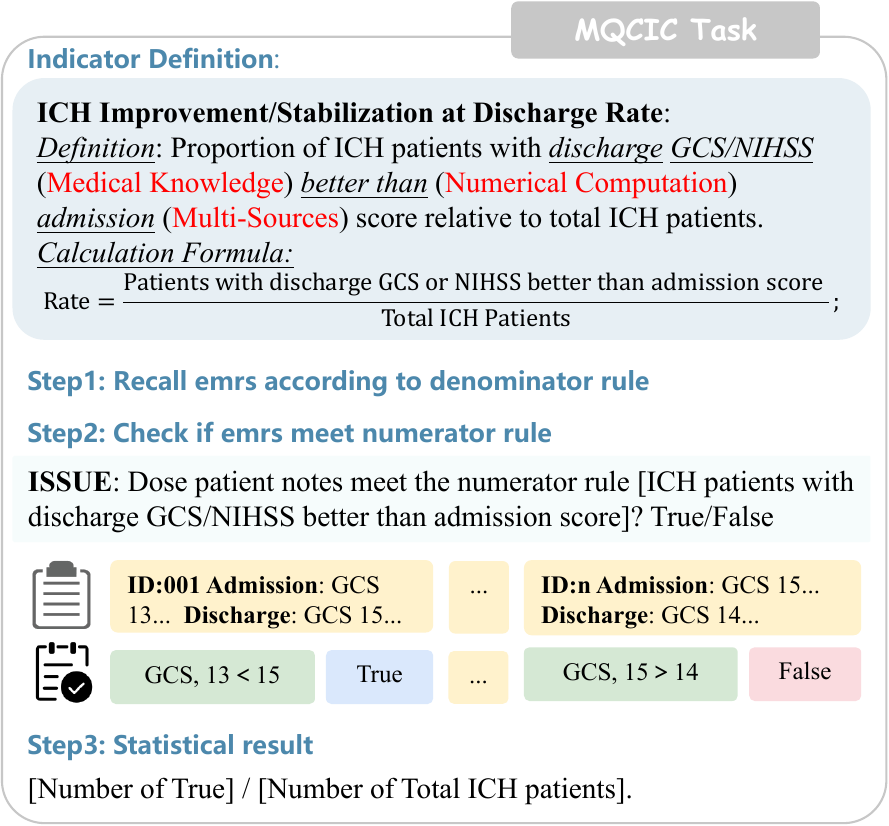}
\caption{An example of calculation progress for ICH improvement/stabilization at discharge rate.
Firstly collect patient records with intracerebral hemorrhage (ICH). Then identify those with discharge scores better than or equal to admission scores. Finally, the proportion of these patients among all ICH cases is calculated.}
\label{fig:task}
\end{figure}
Medical quality control indicators play an essential role in assessing the performance of healthcare institutions~\cite{MQCI,wang2018clinical,example}. 
% For example, the adenoma detection rate and sessile serrated lesion detection rate are critical quality indicators for endoscopists, as they are strongly associated with the risk of post-colonoscopy colorectal cancer and related mortality.~\cite{example}. 
Recently, Large Language Models (LLMs) like GPT-4~\cite{gpt4o} have shown promising capabilities in the medical domain. These include applications such as diagnostic reasoning~\cite{decision}, clinical note generation~\cite{note_generation}, and automated clinical assessment~\cite{assessment}. Such capabilities also prove effective in \textbf{M}edical \textbf{Q}uality \textbf{C}ontrol \textbf{I}ndicator \textbf{C}alculation (MQCIC)~\cite{imqc}.

Traditionally, calculating quality indicators relied on manually constructed rules (regular expressions)~\cite{rule1, rule2}, which is time-consuming~\cite{inefficient}. As illustrated in Figure~\ref{fig:task}, the indicator "ICH Improvement/Stabilization at Discharge Rate" contains <\textit{Definition}, \textit{Calculation Formula}>, which requires \textbf{(i) medical knowledge} regarding the Glasgow Coma Score(GCS) and NIH Stroke Score(NIHSS), reflecting different patient conditions; \textbf{(ii) multiple sources} of information, including both admission and discharge records; and \textbf{(iii) numerical computation or logical reasoning}. With such fine-grained rules, experts develop scripts to identify the relevant data from the unstructed text. However, these quality control indicators related to various diseases are continually refined and expanded over time. Relying solely on fixed scripts or NLP extraction methods is inadequate and lacks generalizability~\cite{nlp1, lack}. 
 
Fortunately, LLMs have demonstrated excellent performance in the transformation as well as decomposition of rules~\cite{plan,plan_train,faithful}. However, several obstacles remain in developing LLM-based clinical applications~\cite{challenge}, especially MQCIC: (i) LLMs struggle to provide accurate, reliable answers for complex clinical reasoning tasks, especially when using Chain-of-Thought (CoT) reasoning~\cite{cot}. (ii) Concerns over LLMs' reliance on opaque, "black-box" methods for clinical decisions, which may erode user trust. Increasing focus is being placed on improving LLM reasoning to follow explicit logical rules~\cite{chainoflogic, rulestress, rulesurvey}, shifting from context-dependent to transparent, rule-based prompting.
% To address these issues, a shift from intrinsic context-based demonstrations to transparent, rule-based prompting demonstrate a promising solution

Therefore, this work aimed to explore leveraging explicit rules to achieve automated indicator calculation using LLMs based on electronic medical records (EMRs). Firstly, we introduce a real-world task MQCIC, and propose an open-source dataset, CMQCIC-Bench, derived from Chinese EMRs on an online Chinese website. The dataset comprises 785 instances spanning 76 indicators. Each instance consists of a Patient Note, a Question, and an Answer. We also provide detailed annotations of clinical facts and explanations. Due to the ambiguity of existing rules that impairs the effectiveness of LLMs, we propose a semi-automatic method to enhance the rule representation. With these refined rules, we introduce the Clinical Fact-based Inferential Rule reasoning (CF-IR) method that disentangles the two abilities during the inference stage. We conducted extensive experiments on 20 representative LLMs across general and medical domain. The evaluation results demonstrate that CF-IR outperforms the CoT method. Furthermore, we investigated the capabilities of clinical fact verification and inferential rule reasoning. 
% Finally, we fine-tuned the Qwen2.5-3b-instruct model using the CMQCIC-Bench dataset and found that it demonstrated outstanding performance in real-world scenarios.
% with general models like GPT-4o, Qwen~\cite{qwen2}, MiniCPM~\cite{minicpm} and LLaMA~\cite{llama3}, along with Chinese medical large language models including HuatuoGPT2~\cite{huatuogpt} and Apollo~\cite{apollo,apollo1}

In summary, the major contributions are as follows:
\begin{itemize}
    \item We introduce a clinical scenario task \textbf{M}edical \textbf{Q}uality \textbf{C}ontrol \textbf{I}ndicator \textbf{C}alculation and propose CMQCIC-Bench, a Chinese open-source dataset with 785 instances, covering 76 different medical quality control indicators.
    \item We propose a semi-automatic method to enhance the rule representation. Then we propose the \textbf{C}linical \textbf{F}act-based \textbf{I}nferential \textbf{R}ule reasoning (CF-IR) method that disentangles the clinical fact verification and inferential rule reasoning actions. 
    \item We conducted comprehensive experiments on 20 representative LLMs, where CF-IR improved performance by 0.43\% in the zero-shot setting and 1.45\% in the one-shot setting.
    \item We analyze errors and explore clinical fact verification and rule reasoning, offering insights to improve MQCIC performance.
    % \item We fine-tune a lightweight LLM based on Qwen2.5-3B, achieving a zero-shot performance score of 71.1—an improvement of 3.3\% compared to Qwen2.5-7B.
\end{itemize}

\section{The Medical Quality Control Indicator Calculation Task}
% In this section, we formally define the input and output of the MQCIC task. Additionally, we formally define the clinical fact verification and inferential rule reasoning capabilities.
Typically, MQCIC involves three steps: (1) Recall relevant EMRs from all cases based on the denominator rules of the indicator. (2) Identify the EMRs that meet the numerator rules from these relevant EMRs. (3) Finally, compute the proportion to determine the indicator's value. The first step can be addressed by matching the ICD-10 codes with diagnostic results. However, the second step is the most challenging, which is the focus of this work. Considering the type of answer is not unique, we define the task as a binary classification problem rather than a cloze task. Thus, the problem is defined as follows: given a Patient Note \(P\) and a Question \(Q\) related to the indicator's rule, the task of MQCIC is to generate the answer \(A=\{True, False\}\). 

% $Input = <Question,\ Patient Note>, Output = <True/False>$, where \textit{Question} represents numerator rules and \textit{Patient Note} represents EMRs that have been filtered according to the denominator rules. 

\section{Dataset Construction}
In this section, we construct a dataset, CMQCIC-Bench, for the MQCIC task. The main content includes the \textbf{data collection} of indicators and patient notes, \textbf{data annotation}, and \textbf{data characteristics}.

\subsection{Data Collection}
We collected indicators and patient notes from two sources. \textbf{Indicators Sources.} We manually curated 76 challenging indicators from authoritative documents\footnote{\href{http://www.ncis.cn/home}{http://www.ncis.cn/home}}, all developed by experts. For each indicator, a rule-related question was constructed for inclusion in the CMQCIC-Bench. \textbf{Patient Notes Sources.} We gathered raw data from a Chinese open-source medical website\footnote{\href{https://www.iiyi.com/}{https://www.iiyi.com/}}. Patient notes meeting the denominator rules were filtered based on ICD-10 codes and diagnostic findings. Finally, we manually removed patient names, hospital information, and other sensitive data to ensure no privacy risks.

\subsection{Data Annotation}
\begin{figure}[t]
\centering
\includegraphics[width=\linewidth]{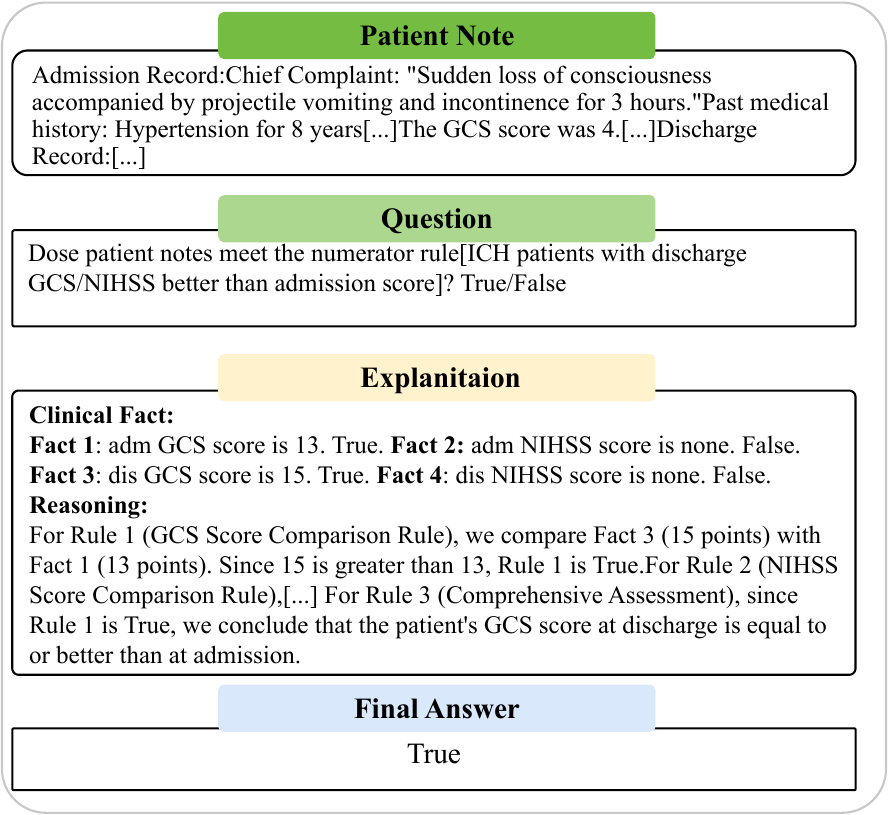}
\caption{Example instance of the CMQCIC-Bench dataset.}
\label{fig:example}
\end{figure}

Specifically, the annotation process uses the following three-step pipeline. \textbf{(1) Clinical fact extraction.} We leverage GPT-4o to extract the original information from EMRs without any modification, then reason based on the context to verify the clinical fact. The clinical facts contain GCS scores, lab exams, medications, etc. \textbf{(2) Answer and explanation generation.} Given the obtained facts, for each instance, we leverage GPT-4o to generate the step-by-step explanation through logical reasoning and a final answer \(\{True, False\}\). \textbf{(3) Data quality control.} Finally, with the guidance of medical experts, annotators are required to check the answer in three facets: fact extraction, logical reasoning, and consistency. Fact extraction and logical reasoning verify the accuracy of the first two stages, while consistency ensures alignment within the patient notes to exclude low-quality cases. In the end, we curated \textbf{785} instances for CMQCIC-Bench, as shown in Figure~\ref{fig:example}, each instance consists of a Patient Note \(P\),  a Question \(Q\), a step-by-step \(Explanation\), and the final answer \(A\). With the same process, we constructed a  CMQCIC-Private dataset derived from patient notes of top-tier tertiary hospitals in China. Ethics committees and experts have rigorously de-identified these data to ensure no privacy leakage risk.

% Additionally, there are situations where facts cannot be determined due to insufficient patient notes, and as a result, no conclusions can be drawn. Since our focus is solely on counting the number of \(True\) labels, we treat any case where a conclusion cannot be reached as \(False\). For example, if the GCS or NIHSS score at admission is unavailable, we cannot determine whether the patient improved by discharge and must assume the requirements are not met. (This error is typically due to incomplete medical documentation and is not addressed here.)

\subsection{Data Characteristics}

\begin{figure}[t]
\centering
\includegraphics[width=\linewidth]{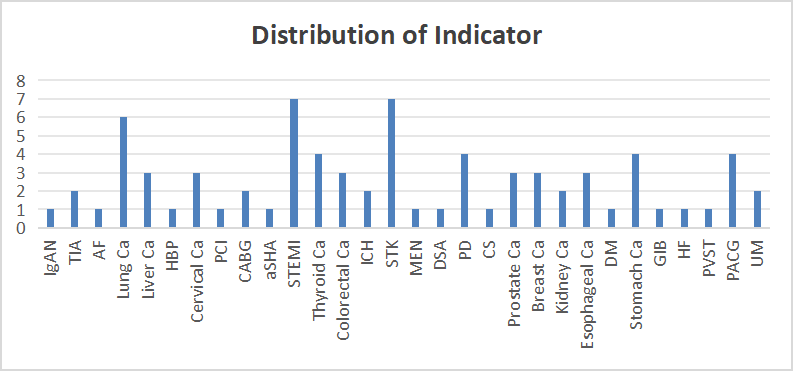}
\caption{The distribution of indicator in CMQCIC-Bench dataset.}
\label{fig:distribution_indicator}
\end{figure}

As shown in Table~\ref{dataset}, we use Tiktoken\footnote{\href{https://github.com/openai/tiktoken}{https://github.com/openai/tiktoken}} to measure sample lengths, yielding average lengths of 380.41 and 520.71, respectively. The shorter average length of each \(P\) in CMQCIC-Bench compared to the private dataset stems from the summarized nature of the source data (e.g., lab exams include only key findings). Despite this, the number of facts ranges from 1 to 13, with averages of 3.59 and 4.02, underscoring the task's demand for multi-step reasoning, consistent with real-world scenarios. Additionally, Figure~\ref{fig:distribution_indicator} illustrates the indicator distribution, which spans 30 diseases.
% and the corresponding data is presented in Figure~\ref{fig:distribution_data}. 
\begin{table}
\centering  
\small  % 或者使用 \scriptsize
\begin{tabular}{c|c|c}\toprule
               & CMQCIC-Bench & CMQCIC-Private \\ \hline 
Indicators     & 76      & 42         \\ \hline 
Instance       & 785     & 314        \\ \hline 
Avg. L of Note & 380.41  & 520.71     \\ \hline 
Avg. L of Q.    & 99.72   & 113.04     \\ \hline 
Min Facts      & 1       & 1          \\ \hline 
Max Facts      & 13      & 13         \\ \hline 
Avg. Facts     & 3.59    & 4.02      
 \\ \hline
 Open-source& Yes&No\\ \hline
 \end{tabular}
 \caption{Statistics of CMQCIC-Bench and CMQCIC-Private datasets. Avg.: average; Q.: question. }
    \label{dataset}
\end{table}

\begin{figure*}[t]
\centering
\includegraphics[width=\linewidth]{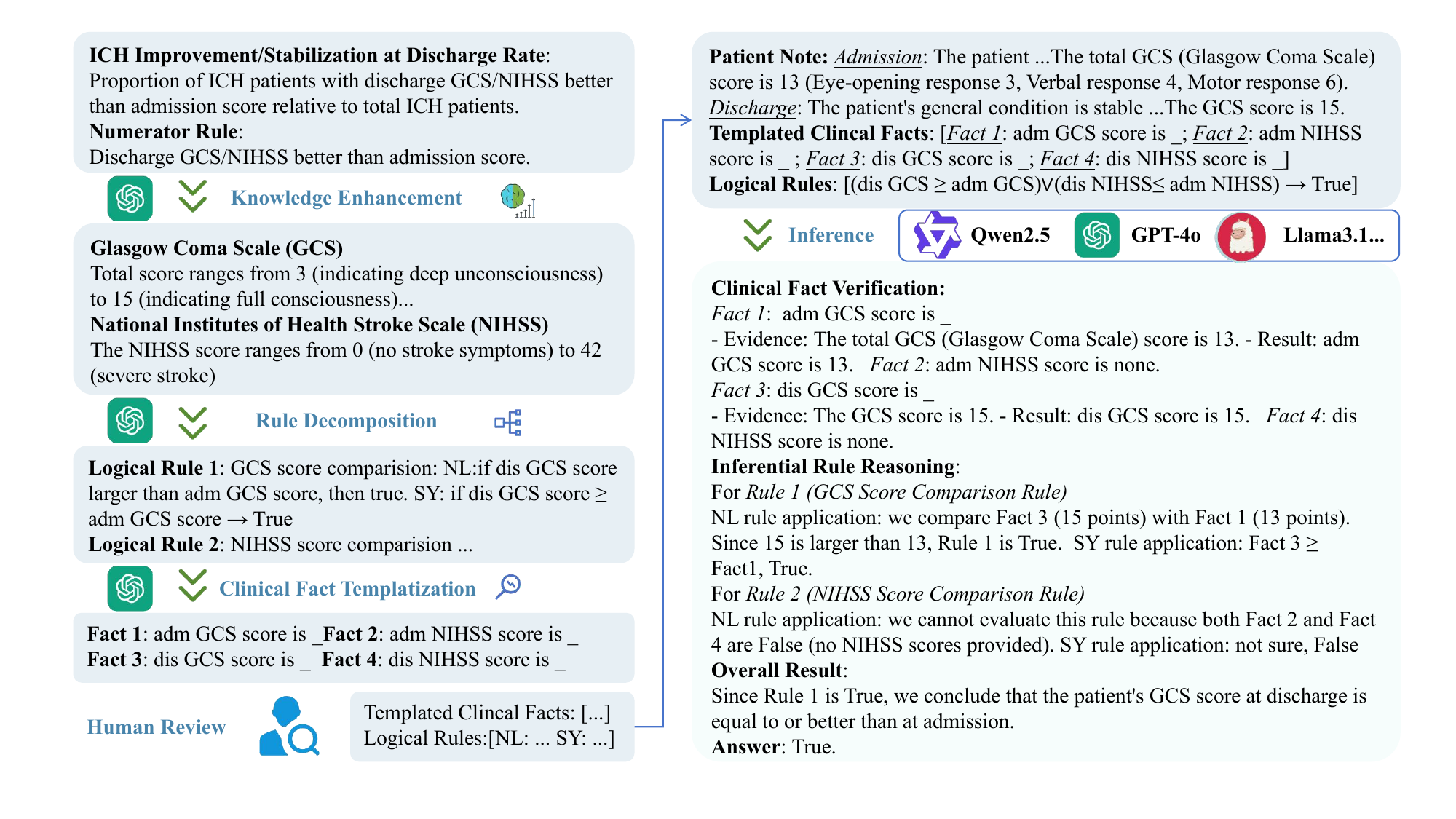}
\caption{An example overview of our method. On the left is the process of Rule Representation Enhancement, where human experts verify each result. On the right is an example illustrating the CF-IR method.}
\label{fig:framework}
\end{figure*}
\section{Method}
As shown in Figure~\ref{fig:task}, the indicators constructed by experts are quite vague, and the underlying medical knowledge can directly affect the implementation of the rules. Therefore,
we propose a semi-automatic method that decomposes them into transparent, templated clinical facts and logical rules. We then introduce the Clinical Fact-based Inferential Rule (CF-IR) reasoning pattern for inference. As illustrated in Figure~\ref{fig:framework}, the method comprises two key components: Rule Representation Enhancement and CF-IR. Additionally, we introduce Automatic CF-IR (ACF-IR) to explore the automatic performance of rule representation enhancement.

% \begin{figure}[t]
% \centering
% \includegraphics[width=\linewidth]{Indicator Conversion.pdf}
% \caption{The overview of Indicator Conversion Process}
% \label{fig:Conversion}
% \end{figure}

% While current research suggests that models like GPT-4 struggle with clinical tasks, they perform well in knowledge-based tasks such as medical exams~\cite{GPT4_medical}. Therefore,
\subsection{Rule Representation Enhancement}
\begin{itemize}
    \item \textbf{Step 1}: \underline{Knowledge Enhancement}. We leverage GPT-4o to recall relevant information instead of collecting additional medical books and guidelines. It aims to resolve rule ambiguities arising from the lack of knowledge. For example, the high and low scores of GCS and NIHSS have different meanings, and splitting the rules based solely on semantic information could result in incorrect logical rules.
    \item \textbf{Step 2}: \underline{Rule Decomposition}.
    We use GPT-4o to break down complex rules into simpler logical rules in both natural and symbolic language. Symbolic language streamlines natural language by converting it into variables, combined with mathematical and logical symbols to build logical expressions. 
    % For example, since the values of the two scores are inversely related to the patient's condition, the complex rule 'discharge GCS/NIHSS better than admission score' is divided into two sub-logical rules to evaluate each score individually.
    % Similar to Chain-of-Logic~\cite{chain-of-logic},
   
    \item \textbf{Step 3}: \underline{Clinical Fact Templatization}. GPT-4o further extracts the clinical facts involved in the logical rules, which are independent. Each clinical fact with a supposed answer set, such as True/False, numerical value unit, etc.
\end{itemize}
In the end, we enlisted human experts to review the enhanced rules for the 76 indicators. 
% While the performance of GPT-4 is already impressive, we still hope that the enhanced rules will achieve the optimal outcome. Therefore, we enlisted human experts to review the enhanced rules for the 76 indicators. 
% We also explore the Automatic performance in Section~\ref{sec:main}.
% We believe that disentangling clinical fact verification and reasoning on facts with explicit logical rules can strengthen the transparency of LLMs.

\subsection{Clinical Fact-based Inferential Rule Reasoning}
Motivated by~\citet{distengtangle}, the model performs two steps during inference: \textbf{Clinical Fact Verification} and \textbf{Inferential Rule Reasoning}. LLM first extracts and verifies the related information using the templated clinical fact. Then, LLM reasoning on the verified clinical facts with explicit logical rules. We believe that disentangling these two distinct abilities improves performance during the inference process and enhances interpretability. Here are some discussions about the two abilities, and we explore deeply in Section~\ref{sec:ability analysis}.

\textbf{(1) Clinical Fact Verification.} 
% Referred to as the 'extraction of relevant patient attributes '~\cite {medcal}. However, when provided with a specific set of criteria, the 'attribute extraction' process resembles more of a 'clinical fact verification'.
Before engaging in clinical reasoning, it is crucial to obtain accurate clinical information\cite{DiReCT}. However, extracting clinical facts and verifying them through reasoning in a long context with noises of over 380.41 tokens is quite challenging. This includes identifying synonyms, linking symptoms to facts (e.g., high GCS indicates better consciousness), understanding medications (e.g., dual antiplatelet therapy), and analyzing surgical indications, all requiring medical knowledge and clinical reasoning.
\begin{table*}
    \centering  
    \small  % 或者使用 \scriptsize
  \resizebox{\textwidth}{!}{
    \begin{tabular}{c l|c c c|c cc}
    \toprule
 & & \multicolumn{3}{c|}{Zero-Shot}& \multicolumn{3}{c}{One-Shot}\\ \hline  
         & &  Standard&  CoT&  CF-IR& ACF-IR 
&CoT
& CF-IR 
\\ \hline  
 % \multicolumn{6}{c}{General Models}\\ \hline  
\multirow{14}{*}{\rotatebox{90}{General}}&
MiniCPM3-4B~\cite{minicpm}
&    63.31 &    \cellcolor[HTML]{E6FFE6}72.10 &    68.91 &    78.98 &   \cellcolor[HTML]{E6FFE6}83.56 &    82.67 
\\ \cline{2-8}
 & Internlm2.5-1.8B~\cite{internlm2}& 56.17& \cellcolor[HTML]{E6FFE6}56.18& 54.14& 65.85& \cellcolor[HTML]{E6FFE6}68.91&64.07
\\
 & Internlm2.5-7B~\cite{internlm2}& 63.31& 73.12&\cellcolor[HTML]{E6FFE6}79.49& 77.07& 84.07&\cellcolor[HTML]{E6FFE6}84.45\\
 & Internlm2.5-20B~\cite{internlm2}& 69.04& 77.57& \cellcolor[HTML]{E6FFE6}80.63& 83.06& 86.36&\cellcolor[HTML]{E6FFE6}88.78\\   \cline{2-8}
 &Qwen2.5-0.5B~\cite{qwen2}&    54.26 &    \cellcolor[HTML]{E6FFE6}56.05 &    53.88 &    52.86 &   \cellcolor[HTML]{E6FFE6}61.01 &    53.63 
\\   
 &Qwen2.5-1.5B~\cite{qwen2}&    \cellcolor[HTML]{E6FFE6}66.11 &    63.43 &    62.42 &    60.50 &   71.21 &    \cellcolor[HTML]{E6FFE6}73.24\\  
  &Qwen2.5-3B~\cite{qwen2}&    60.38 &    \cellcolor[HTML]{E6FFE6}73.37 &    67.64 &  79.36$\dagger$&   77.83 &    \cellcolor[HTML]{E6FFE6}82.03\\   
         &Qwen2.5-7B~\cite{qwen2}&     66.49 &    82.80 &     \cellcolor[HTML]{E6FFE6}82.92&     85.73$\dagger$&   85.22 &    \cellcolor[HTML]{E6FFE6}89.93\\   
         &Qwen2.5-14B~\cite{qwen2}&     78.98 &     82.03&    \cellcolor[HTML]{E6FFE6}86.11&     84.96 &   87.89 &    \cellcolor[HTML]{E6FFE6}91.59 
\\
 & Qwen2.5-32B~\cite{qwen2}&    75.54 &    86.49&   \cellcolor[HTML]{E6FFE6}87.21&   92.35$\dagger$&    89.80 &   \cellcolor[HTML]{E6FFE6}94.77 
\\   
         &Qwen2.5-72B~\cite{qwen2}& 87.77 &  87.51 & \cellcolor[HTML]{E6FFE6}92.73 &  91.33$\dagger$&  90.95 & \cellcolor[HTML]{E6FFE6}95.54\\   \cline{2-8}
         &llama3.1-8B~\cite{llama3}
&     48.53 &     63.05 &  \cellcolor[HTML]{E6FFE6}78.34&     73.88 &   81.52 & \cellcolor[HTML]{E6FFE6}85.85\\
&llama3.1-70B~\cite{llama3}
&    82.54 &     \cellcolor[HTML]{E6FFE6}85.85 &     85.47 &     84.45 &   88.53 &    \cellcolor[HTML]{E6FFE6}91.84\\  \cline{2-8}
&GPT-4o~\cite{gpt4o}&     77.45 &  88.91 &    \cellcolor[HTML]{E6FFE6}91.84&     90.57 &  91.59 &  \cellcolor[HTML]{E6FFE6}93.88 
\\ 
         \hline  
         \multirow{6}{*}{\rotatebox{90}{Medical}}  
         &HuatuoGPT2-7B~\cite{huatuogpt}
&     54.01 &     \cellcolor[HTML]{E6FFE6}54.26 &     49.55 &     48.66 &   53.50 &    \cellcolor[HTML]{E6FFE6}56.81 
\\
         &HuatuoGPT2-14B~\cite{huatuogpt}&     53.63 &     \cellcolor[HTML]{E6FFE6}55.28 &     46.36 &     37.19 &   \cellcolor[HTML]{E6FFE6}52.10 &    43.31 
\\ \cline{2-8}
 & Apollo2-0.5B~\cite{apollo}&    39.55 &    41.96 &    41.14 &    57.19$\dagger$&   54.39 &    \cellcolor[HTML]{E6FFE6}65.47 
\\
 & Apollo2-1.5B~\cite{apollo}&    53.31 &    52.03 &    50.82 &    52.61 &   66.11 &    65.22 
\\
         &Apollo2-7B~\cite{apollo}&     57.57 &     60.00 &    \cellcolor[HTML]{E6FFE6}61.91&     48.91 &   \cellcolor[HTML]{E6FFE6}71.71 &    65.35 
\\ 
 &Apollo-72B~\cite{apollo1}&    68.91 &    \cellcolor[HTML]{E6FFE6}76.24 &    72.61 & 
   80.63 &   
   86.11 & 
   \cellcolor[HTML]{E6FFE6}86.36 
\\ \hline  
 \multicolumn{2}{c|}{Average \cellcolor[HTML]{EFEFEF}}& \cellcolor[HTML]{EFEFEF}63.84&\cellcolor[HTML]{EFEFEF} \underline{69.41}& \cellcolor[HTML]{EFEFEF}\textbf{69.71}& \cellcolor[HTML]{EFEFEF}71.31& \cellcolor[HTML]{EFEFEF}\underline{76.62}& \cellcolor[HTML]{EFEFEF}
\textbf{77.73}\\ \hline
\multicolumn{2}{c|}{\cellcolor[HTML]{CBCEFB} Human} & \multicolumn{6}{c}{\cellcolor[HTML]{CBCEFB} \textbf{95.00}} \\ \hline
    \end{tabular}
    }
    \caption{Aggregated performance (micro-average accuracy) across all indicators on CMQCIC-Bench, using general and medical LLMs. \textbf{Bold} denotes the best performance. \underline{Underline} denotes the second performance. \colorbox[HTML]{E6FFE6}{Green} denotes the best performance in certain LLM.  $\dagger$ denotes ACF-IR outperforms the CoT.}
    \label{main_result}
\end{table*}

\textbf{(2) Inferential Rule Reasoning.}
Reasoning can be categorized in multiple ways~\cite{NLsurvey}. In this study, we specifically define 'Inferential Rule Reasoning' as the capability to deduce a final conclusion by applying logical rules to multiple clinical facts. Specifically, for each templated clinical fact, the LLM first performs targeted information extraction and verification. Subsequently, it applies explicit logical rules to conduct structured reasoning based on the verified clinical facts.

\section{Experiments}
\label{sec:experiments}
\definecolor{myred}{RGB}{240, 128, 128}
\definecolor{myblue}{RGB}{70, 130, 180}
\definecolor{mygreen}{RGB}{50, 205, 50}

We now describe the experimental setup, empirically evaluate CF-IR, and compare it with existing methods using the CMQCIC-Bench dataset.
\subsection{Settings}
Here are the details of the models, methods, and evaluation metrics.

\textbf{Models.} We select a representative set of 14 general large language models including GPT-4\footnote{gpt-4o-2024-08-06}~\cite{gpt4o}, Qwen2.5 series~\cite{qwen2}, Internlm2.5 series~\cite{internlm2}, Llama3.1 series~\cite{llama3}, MiniCPM3-4B~\cite{minicpm}, as well as 6 medical large language models, comprising the Apollo series~\cite{apollo,apollo1}, and HuatuoGPT2 series~\cite{huatuogpt}. We omit the "chat" and "instruct" in Table~\ref{main_result}. 

\begin{table*}
    \centering
    \small  % 或者使用 \scriptsize
  % \resizebox{\textwidth}{!}{
    \begin{tabular}{cp{4.4cm}p{4.0cm}p{5.3cm}}
    \toprule 
         Type&  Question&  Patient Note & Error of LLMs\\ \hline 
         A& Can the screening for anxiety and depression symptoms be considered complete based on the medical records of hospitalized Parkinson's disease patients?& On the initial visit and the 34th day after treatment, the patient's H-Y stage decreased from stage 3 to 2.5, \textcolor{mygreen}{the HAMD score dropped from 23 to 2, the Beck Depression Inventory (BDI) score decreased from 14 to 4}, the PDSS-2 score dropped from 10 to 2,[...].  & Based on the provided electronic medical record information, there are no clear mentions of specific results for anxiety and depression symptom screenings. Therefore, we are unable to determine the values for the following facts: \textcolor{red}{Anxiety symptom screening = Not sure Depression symptom screening =  Not sure.} \textcolor{mygreen}{\#\#HAMD and BDI can evaluate the depression.}\\ \hline 
         B&  Please assess if the patient's coronary intervention meets post-operative success criteria: residual stenosis <20\% after stenting or <50\% after PTCA, with a TIMI grade of 3. & Treatment Process: [...] Post-stent implantation, there was \textcolor{mygreen}{no significant residual stenosis at the LAD lesion and TIMI grade II}. & Based on the provided instructions and medical records, [...]  After stent implantation, there was no significant residual stenosis in the patient's left coronary artery LAD. Therefore, the answer is "Yes" \textcolor{mygreen}{\#\#comment: TIMI grade must be III.}  \\ \hline 
    \end{tabular} 
    \caption{Examples of two main error types. \textcolor{mygreen}{evidence}, \textcolor{red}{incorrect rationale}, and \textcolor{mygreen}{\#\# comment} is colored.}
    \label{tab:error_example}
\end{table*}

\textbf{Methods.} We assess six prompt-based methodologies: (1) Standard Prompt, utilizing solely the original rules and patient notes; (2) Zero-Shot CoT~\cite{zeroshotcot}, enhanced with the directive "Let’s think step by step"; (3) For CF-IR prompt, we leverage human-reviewd fact templates and logical rules to derive the answer; (4) Specifically, we examine the \textbf{One-shot CoT}~\cite{cot}; (5) To explore the ability of LLM for rule representation enhancement, we introduce \textbf{ACF-IR}, an automated framework that enables LLMs to decompose rules and then conduct CF-IR; (6) Additonally, we set \textbf{One-Shot CF-IR}. 

For each indicator, we selected an example outside of CMQCIC-Bench. The outputs for these examples were generated by GPT-4o in a zero-shot setting and were carefully reviewed and annotated by human evaluators.

\textbf{Evaluation.} Following prior works~\cite{chain-of-logic, rulestress}, the final answer for each instance was labeled as '\textit{True}/\textit{Yes}' or '\textit{False}/\textit{No}', enabling us to use accuracy as the \textbf{outcome evaluation} metric. To evaluate \textbf{step-wise reasoning}, we compared responses against ground truth using DeepSeek~\cite{gpteval}, assessing two dimensions: \textbf{Fact Faithfulness} (relevance to the original text) and \textbf{Fact Correctness} (accuracy of the fact result). Scores range from 0 (irrelevant/incorrect) to 1 (fully relevant/correct). We utilized DeepSeek~\footnote{\href{https://www.deepseek.com/}{https://www.deepseek.com/}} to extract facts from gold explanations and score model responses accordingly. Scores were averaged per instance to accommodate varying fact counts, resulting in an overall average score. Below is a formal definition of the metrics:
\begin{equation}
    FC_{i} = \frac{\sum_{j}^{m}Judge(fact_{j}, r)}{m},
    % avg\_fact correctness = \frac{\sum_{i}^{n}correctness_{i}}{n}
\end{equation}

\begin{equation}
    FF_{i} = \frac{\sum_{j}^{m}Judge(fact_{j},r)}{m},
 % avg\_fact faithfulness = \frac{\sum_{i}^{n}faithfulness_{i}}{n}
\end{equation}
where Judge($\cdot$) represents the LLMs, outputting 0 or 1. The m denotes the number of facts in the i-th instance. The $fact_{j}$ denotes the j-th fact of the i-th instance. \textbf{Human evaluation}, we designed regex to extract key information, subsequently assessed by experts.

\textbf{Implement Details.} We conduct all experiments on H800 and use VLLM~\footnote{\href{https://github.com/vllm-project/vllm}{https://github.com/vllm-project/vllm}} to accelerate for general LLMs. Specifically, we load the medical LLMs directly. Additionally, we set the max\_new\_tokens = 1024; repetition\_penalty = 1.2; temperature = 0.001. The experiments were run three times with random seeds, and the scores were averaged.

\subsection{Main Results}
\label{sec:main}

Table~\ref{main_result} presents our evaluation results of various LLMs on the CMQCIC-Bench dataset.

\textbf{(1) Current leading general LLMs perform better than medical LLMs.} Qwen2.5-32B/72B-Instruct, and GPT-4o score similarly at 94.77, 95.54, and 93.88, respectively, while medical LLMs lag, with Apollo-72b scoring only 86.36. Only Qwen2.5-72B-Instruct nears human performance, highlighting the ongoing challenge of the MQCIC task for current methods and LLMs.

% This conclusion aligns with findings from other evaluations~\cite{CliMedBench, medcal}, which attribute the subpar performance to limitations in language understanding and instruction-following capabilities.

\textbf{(2) CF-IR methods perform better than CoT across different parameters and models.} In zero-shot and one-shot settings, the average score of CF-IR improves by 0.43\% and 1.45\%, respectively, compared to CoT. Unlike the CoT method, which performs reasoning along random paths, our approach integrates explicit logical rules with verifiable facts, enhancing the stability and interpretability of LLMs. While CF-IR demonstrated strong performance across various parameters in the one-shot setting, we observed that in the zero-shot scenario, CF-IR outperformed CoT only on general models with parameters $\geq$ 7B. We will analyze our improvement in Section~\ref{sec:analysis}.

\textbf{(3) One-Shot setting can bring significant improvements.} In general, after providing the examples, CoT and CF-IR achieved improvements of 10.38\% and 11.50\%, respectively, the performance of all models showed significant improvements in the one-shot setting for both the CoT and CF-IR methods except HuatuoGPT2. This may stem from HuatuoGPT's fine-tuning data being predominantly centered around QA tasks~\cite{huatuogpt}, without incorporating clinical scenarios, and weak in instruction-following.
% We manually reviewed HuatuoGPT2's responses and found that the examples provided little benefit. This may stem from HuatuoGPT's fine-tuning data being predominantly centered around QA tasks~\cite{huatuogpt}, without incorporating clinical scenarios, and weak in instruction-following. Consequently, its one-shot performance lags behind that of the Apollo series models.

\textbf{(4) Automated rule representation enhancement remains challenging.} While CF-IR achieves strong performance (\textbf{77.73}) with enhanced rule representation, ACF-IR's automated approach scores lower (71.31), underperforming CoT. Notably, only Apollo2-0.5B and specific Qwen2.5 variants (3B, 7B, 32B, 72B) surpass CoT in one-shot settings, revealing the limitations of intrinsic model planning capabilities~\cite{chain-of-logic, BoT}. A promising direction involves leveraging advanced open-source models (e.g., GPT-4o) or specialized plan training~\cite{plan_train} for rule decomposition, complemented by medical models for inference.

\section{Empirical Analysis and Discussion}
\label{sec:analysis}
In this section, we analyze errors and evaluate step-wise reasoning, further exploring clinical fact verification and inferential rule reasoning capabilities.

\subsection{Error Analysis}
Firstly, we categorize errors into three types: Type A, B and C, representing errors in clinical fact verification, inferential rule reasoning, and other types, respectively. We display the example of two main error types in CMQCIC-Bench in Table~\ref{tab:error_example}.

Building on prior work~\cite{medcal}, we employ DeepSeek to classify error types by comparing LLM outputs with ground truth in CMQCIC-Bench, facilitating a granular error analysis across LLMs. Since incorrect clinical facts can propagate and affect inferential rule reasoning, we focus on identifying the earliest error type. A manual review of 200 randomly sampled DeepSeek-annotated errors confirmed an 87\% accuracy, validating our approach for analyzing error types in all CF-IR responses. As shown in Table~\ref{average_error}, providing demonstrations reduces Type A and B errors, highlighting the value of exemplars. While CF-IR does not mitigate clinical fact verification errors, it significantly improves reasoning accuracy due to its structured logical framework. Further details are available in Appendix~\ref{apendix:error} (Tables~\ref{cot_error} and~\ref{CF-IR_error}).
\begin{table}
    \centering  
      % 或者使用 \scriptsize
    % \small
    \begin{tabular}{c|c c| cc}
    \toprule
  & \multicolumn{2}{c|}{Zero-Shot}&\multicolumn{2}{c}{One-Shot}\\ \cline{1-5}  
          Error Type&  CoT&  CF-IR& CoT& CF-IR\\ \cline{1-5}
   clinical fact & 0.19& 0.23& 0.17$\textcolor{blue}{\downarrow}$& 0.17$\textcolor{blue}{\downarrow}$\\ 
reasoning&    0.11&    0.07& 0.07$\textcolor{blue}{\downarrow}$& 0.05$\textcolor{blue}{\downarrow}$\\  
 other& 0.00& 0.01& 0.00-& 0.00$\textcolor{blue}{\downarrow}$\\
 Total& 0.31& 0.30& 0.23$\textcolor{blue}{\downarrow}$& 0.22$\textcolor{blue}{\downarrow}$\\ \hline
    \end{tabular}
    \caption{Error type distribution of LLMs on CMQCIC-Bench. Arrows represent the changes from zero-shot to one-shot. We averaged all the models' performances.}
    \label{average_error}
\end{table}

\subsection{Evaluation on Step-Wise Reasoning}
As shown in Table~\ref{tab:step-wise eval}, the step-aware evaluation metrics decreased by 14.38 and 17.19 points, respectively, compared to the outcome evaluation results. This suggests that the model often makes clinical fact verification errors during the reasoning process, even when the final result is correct.
\begin{table}
    \centering  
      % 或者使用 \scriptsize
    \small
\setlength{\tabcolsep}{1.3mm}
\begin{tabular}{ccccc}
\toprule
\textbf{Models}         & \textbf{Methods}         & \textbf{FC} & \textbf{FF} & \textbf{ACC} \\ \hline
\multirow{4}{*}{Qwen2.5-72b}    & zero-shot CoT           & 68.09                    & 68.07                      & 87.51             \\ 
                        & zero-shot CF-IR         & 76.34                    & 76.83                      & 92.73             \\ 
                        & one-shot CoT            & 69.26                    & 69.92                      & 90.95             \\ 
                        & one-shot CF-IR          & 90.45                    & 86.20                      & 95.54             \\ \hline
\multirow{4}{*}{Qwen2.5-32b}    & zero-shot CoT           & 66.63                    & 66.42                      & 86.49             \\ 
                        & zero-shot CF-IR         & 72.86                    & 71.53                      & 87.21             \\ 
                        & one-shot CoT            & 69.73                    & 70.44                      & 89.80             \\ 
                        & one-shot CF-IR          & 84.61                    & 76.78                      & 94.77             \\ \hline
\multirow{4}{*}{Qwen2.5-14b}    & zero-shot CoT           & 68.25& 66.44& 82.03 
\\ 
                        & zero-shot CF-IR         & 70.87& 65.14& 86.11
\\ 
                        & one-shot CoT            & 71.32& 69.41& 87.89
\\ 
                        & one-shot CF-IR          & 83.48& 78.22& 91.59
\\ \hline
\multirow{4}{*}{Qwen2.5-7b}& zero-shot CoT           & 67.31& 63.36& 82.80 
\\ 
                        & zero-shot CF-IR         & 67.03& 65.76& 82.92
\\ 
                        & one-shot CoT            & 66.60& 66.45& 85.22
\\ 
                        & one-shot CF-IR          & 77.97& 71.88& 89.93
\\ \hline
\multirow{4}{*}{llama3.1-70b}   & zero-shot CoT           & 65.24& 62.49                      & 85.85             \\ 
                        & zero-shot CF-IR         & 71.70                    & 65.23                      & 85.47             \\ 
                        & one-shot CoT            & 69.41                    & 68.59                      & 88.53             \\ 
                        & one-shot CF-IR          & 83.15                    & 77.50                      & 91.84             \\ \hline
\multirow{4}{*}{llama3.1-8b}    & zero-shot CoT           & 57.02                    & 57.52                      & 63.05             \\ 
                        & zero-shot CF-IR         & 64.89                    & 61.25                      & 78.34             \\ 
                        & one-shot CoT            & 68.13                    & 65.61                      & 81.52             \\ 
                        & one-shot CF-IR          & 78.40                    & 70.38                      & 85.85             \\ \hline
\multicolumn{2}{c}{\textbf{Average}}& 72.03& 69.22& 86.41\\ \bottomrule
\end{tabular}
\caption{Comparison of step-wise and outcome evaluation. \textbf{FC} denotes Fact Correctness. \textbf{FF} denotes Fact Faithfulness. The results of \textbf{ACC} sourced from Table~\ref{main_result}.}
\label{tab:step-wise eval}
\end{table}

\begin{figure}[t]
\centering
\begin{subfigure}[b]{0.95\linewidth}
    \includegraphics[width=\linewidth]{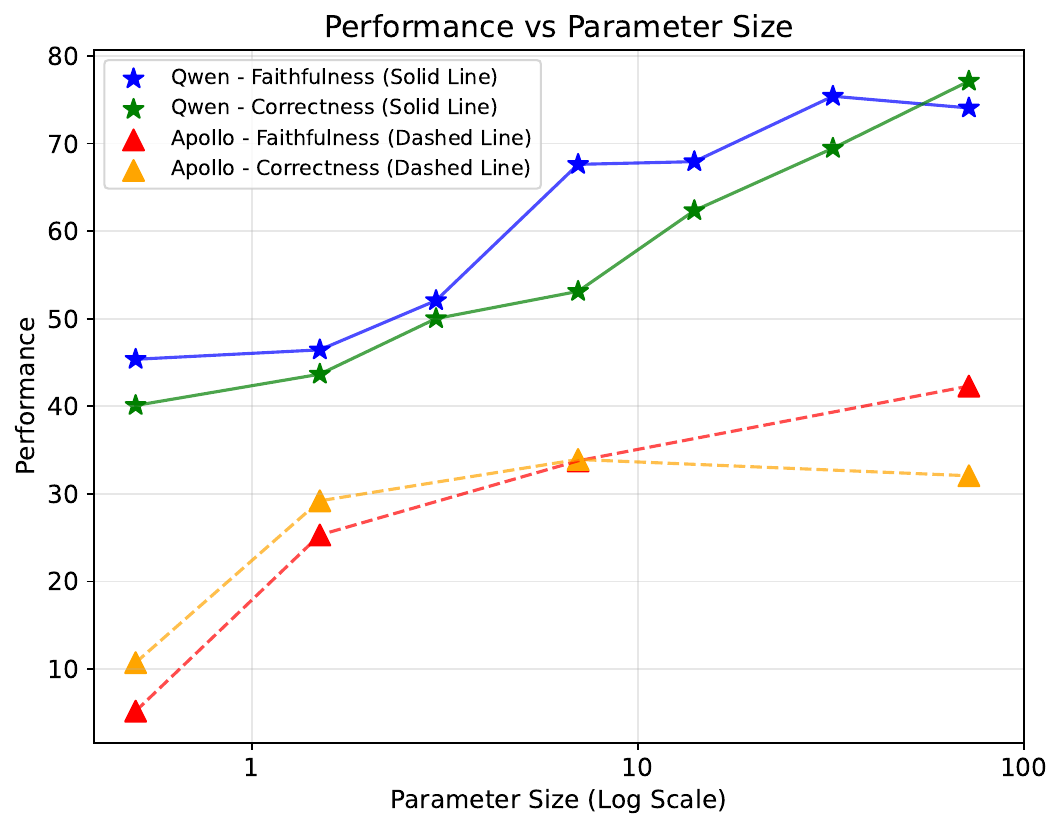}
    % \caption{Clinical fact verification ability of Qwen and Apollo series on CMQCIC-Bench.}
    % \label{fig:FV}
\end{subfigure}
\hfill
\begin{subfigure}[b]{0.95\linewidth}
    \includegraphics[width=\linewidth]{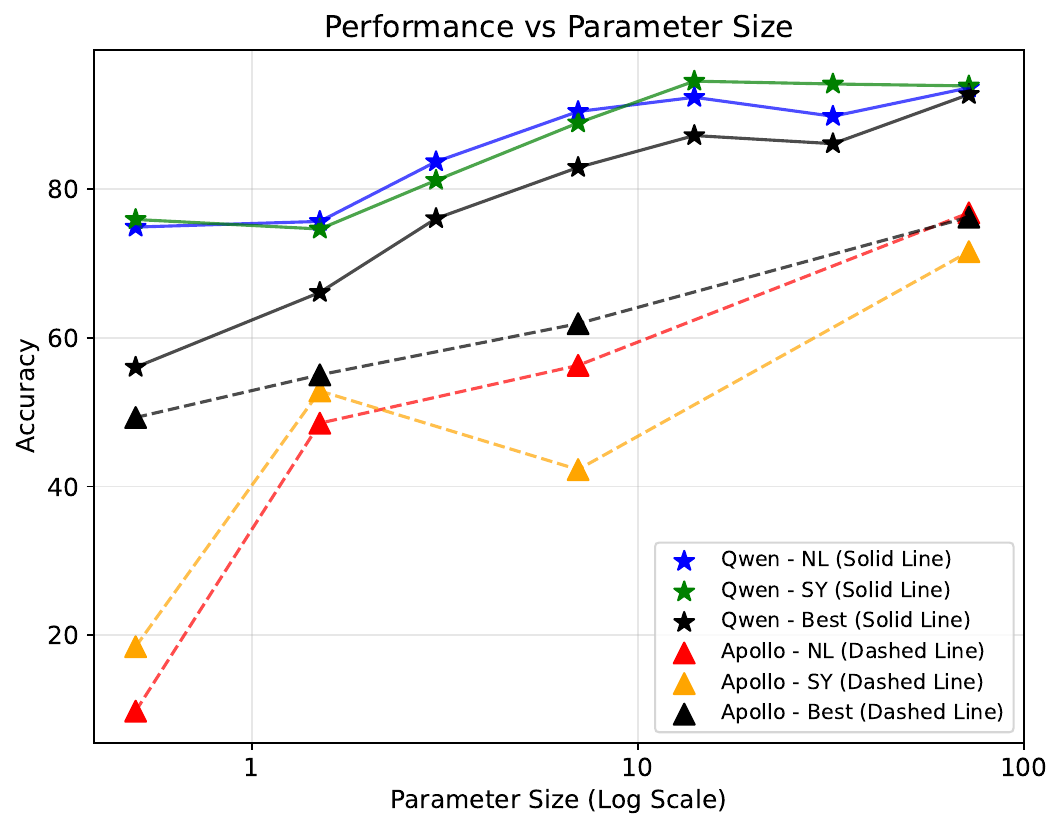}
    % \caption{Inferential rule reasoning ability of Qwen and Apollo series on CMQCIC-Bench. NL denotes natural language; SY denotes symbolic language.}
    % \label{fig:IR}
\end{subfigure}
\caption{Clinical fact verification and inferential rule reasoning abilities of Qwen and Apollo series on CMQCIC-Bench. NL denotes natural language; SY denotes symbolic language.}
\label{fig:combined}
\end{figure}

\subsection{Analysis on Clinical Fact Verification and Inferential Rule Abilities}
\label{sec:ability analysis}
While the CF-IR method enhances inference performance, we further investigate its two core capabilities: \textbf{Clinical Fact Verification} and \textbf{Inferential Rule Reasoning}. For \textbf{Clinical Fact Verification}, we define the input as <\textit{Patient Note}, \textit{Templated Clinical Fact}, \textit{Question}> and the output as <\textit{Reasoning}, \textit{Final Answer}>, evaluated using \textbf{Fact Faithfulness} and \textbf{Fact Correctness}. Unlike step-wise reasoning, we test each fact independently to avoid contextual interference. For \textbf{Inferential Rule Reasoning}, providing verified facts as input to minimize errors, the input is <\textit{Verified Clinical Facts}, \textit{Logical Rules}>, and the output is <\textit{Explanation}, \textit{Final Answer}>, evaluated using labels like '\textit{True}/\textit{Yes}' or '\textit{False}/\textit{No}' for both natural (\textbf{NL}) and symbolic (\textbf{SY}) languages. All experiments are conducted in a zero-shot setting. Additional results are available in Table~\ref{ablation} in Appendix~\ref{appendix:FV}.

% \begin{figure}[t]
% \centering
% \includegraphics[width=0.5\linewidth]{fv.pdf}
% \caption{Clinical fact verification ability of Qwen and Apollo series on CMQCIC-Bench.}
% \label{fig:FV}
% \end{figure}
% \begin{figure}[t]
% \centering
% \includegraphics[width=0.5\linewidth]{ir.pdf}
% \caption{Inferential rule reasoning ability of Qwen and Apollo series on CMQCIC-Bench. NL denotes natural language; SY denotes symbolic language.}
% \label{fig:IR}
% \end{figure}

As shown in Figures~\ref{fig:combined}, we find that: \textbf{(1)} Both Qwen and Apollo exhibit performance-scale correlations across capabilities. \textbf{(2)} Fact verification performance significantly declines, consistent with Table~\ref{tab:step-wise eval}. \textbf{(3)} For inferential reasoning, Qwen performs comparably in natural and symbolic settings, while Apollo shows stronger natural language robustness. \textbf{(4)} With correct facts, Qwen surpasses previous best results (standard, CoT, CF-IR) in zero-shot settings, whereas Apollo underperforms, likely due to Qwen's extensive logical reasoning training. See Appendix~\ref{appendix:FV} for additional results.
% This could be because Qwen has been trained on more logical reasoning data than Apollo, which supports Qwen to perform better with true clinical facts.

% In the end, we propose that 7B represents a critical turning point for two key capabilities, which directly influence the performance of CF-IR in the zero-shot setting on CMQCIC.

\subsection{Benefit of fine-tuning}
\begin{figure}[t]
\centering
\includegraphics[width=0.95\linewidth]{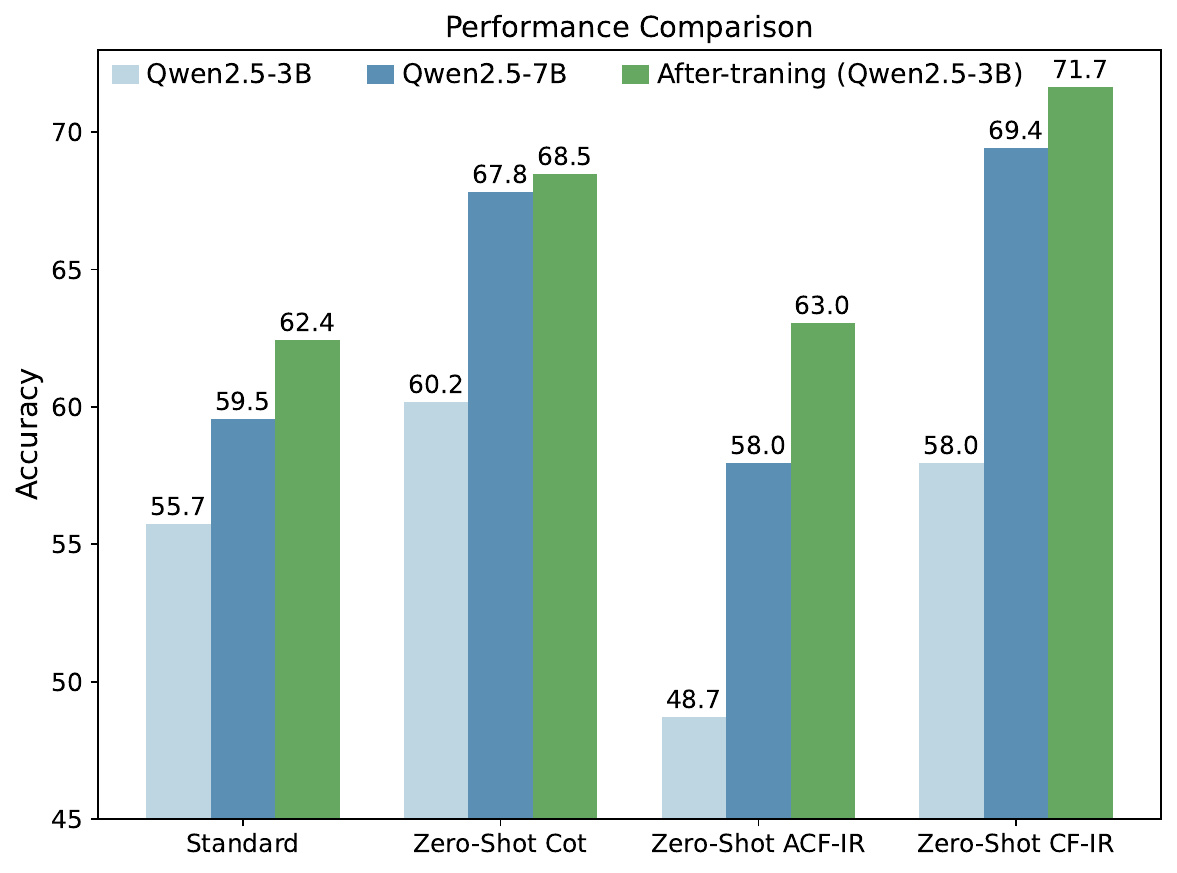}
\caption{Analysis of fine-tuning benefit. Performance of three models on the CMQCIC-Private dataset.}
\label{fig:train}
\end{figure}

We fine-tuned the Qwen2.5-3B-Instruct model using LoRA~\cite{lora} for 3 epochs on the CMQCIC-Bench, with evaluation on CMQCIC-Private. The data format follows ACF-IR: Input: <\textit{Instruction}, \textit{Patient Note}, \textit{Rule}>; Output: <\textit{Knowledge}, \textit{Templated Clinical Facts}, \textit{Logical Rules}, \textit{Clinical Fact Verification}, \textit{Inferential Rule Reasoning}, \textit{Final Answer}>. As depicted in Figure~\ref{fig:train}, the fine-tuned 3B model achieves comparable or superior performance to the 7B model in real-world scenarios. In zero-shot settings, it demonstrates significant improvements across all methods, with gains of 6.7, 8.3, 14.3, and 13.7, confirming the feasibility of distillation rule enhancement for smaller models. More details are provided in Appendix~\ref{appendix:training}.
% Notably, the improvement in Auto CF-IR was particularly pronounced. This enhancement is likely due to similar tasks within the CMQCIC-Bench dataset.

\section{Related Work}
\subsection{Rule-based LLM reasoning}
While reasoning demonstrated a fundamental capability of LLM on applications~\cite{li2024fundamental}, there are many research such as CoT~\cite{cot}, CoT-sc~\cite{cot-sc}, ToT~\cite{ToT}, etc. However, there are more attention on rule-enhanced methods~\cite{LLMrule1, LLMrule2, LLMrule3}. Reasoning based on facts and deriving answers from logical rules is referred to as inferential rule following ability~\cite{rulesurvey}. Leveraging such ability that integrating explicit rules with LLMs has gained significant attention. For instance, ~\citet{chain-of-logic} utilized the IRAC framework to tackle legal tasks with LLMs, emphasizing the application of legal rules. Additionally, ~\citet{ruleapplication} proposed a neurosymbolic framework for multi-step rule application. Despite the current limitations of LLMs in rule-based reasoning~\cite{rulestress}, our work demonstrates that such rule-based reasoning outperforms CoT reasoning in the MQCIC task.

\subsection{LLM Evaluations in Clinical Scenarios}
While LLMs have shown impressive capabilities in medical knowledge recall and reading comprehension on medical exams~\cite{GPT4_medical, M-QALM}, their effectiveness in real-world clinical applications remains a critical area of evaluation. For example, ~\citet{CliMedBench} assesses LLMs across 14 expert-curated clinical scenarios, including diagnosis, discharge summaries, and medical consultations. Similarly, ~\citet{medcal} introduces MedCal-Bench, a benchmark designed to evaluate inferential rule reasoning in medical contexts, while ~\citet{Hou} simulates a multi-step diagnostic process to test clinical reasoning capabilities. Furthermore, ~\citet{factdecomp,verifact} explore LLMs' abilities in clinical fact decomposition and verification. In this work, we focus on evaluating LLMs in the MQCIC task, with a specific emphasis on their performance in clinical fact verification and inferential rule reasoning, providing a detailed analysis of these two critical abilities.

% (2) \textbf{Insufficient hardware computing power to support large-parameter models.} Previous work has explored the upper limits of LLMs in clinical tasks, while some research focuses on training cost-effective models for real-world applications~\cite{apollo, apollo1, NC}. 
% In line with these efforts, we also report the performance of lightweight models and our training results, aiming to explore a more balanced trade-off approach.

\section{Conclusion}
In this work, we present MQCIC, a novel task, and CMQCIC-Bench, an open-source dataset derived from Chinese EMRs. We propose a semi-automatic approach to refine rule representation and introduce CF-IR, a disentangled inference method. Experimental results show that CF-IR surpasses CoT in performance. Error analysis reveals enhanced capabilities in clinical fact verification and inferential rule reasoning. Additionally, we evaluate step-wise reasoning and conduct a detailed investigation of the two abilities. Our work aims to advance the application of LLMs in MQCIC tasks and offers deeper insights into these essential capabilities.

\section*{Limitations and Future Work}
While we construct a CMQCIC-Bench dataset and evaluate LLMs' clinical fact verification and inferential rule reasoning abilities, several limitations can be improved. (1) Due to the difficulty of manually verifying each sample, our dataset only contains 785 instances. (2) We have only located a comprehensive Chinese document on medical quality control indicators. As a result, our dataset consists solely of Chinese EMRs, and we are also leaning toward selecting Chinese LLMs for our analysis. (3) While we observed a significant improvement in model performance with the one-shot demonstration, benchmarking the model with few-shot instances could have further enhanced accuracy, a scenario we did not test. (4) Although we propose the CF-IR method, which performs well across various LLMs with an enhanced rule representation reviewed by humans, decomposing the rules with a smaller LLM that lacks strong planning capabilities remains a challenge.

\section*{Ethical Consideration}
The medical cases are sourced from the iiyi website, where doctors voluntarily contribute and share their information. The data is explicitly authorized for use in research and educational activities. To safeguard patient privacy, our dataset excludes any personally identifiable details, such as patient names, hospital information, or other sensitive data. As a result, there is no risk of privacy violations related to our dataset. Furthermore, all data usage adheres to ethical guidelines and regulations governing medical information and research.
\section*{Acknowledgments}
This paper was supported by the National Natural Science Foundation of China (No. 62306112), Shanghai Sailing Program (No. 23YF1409400), and Shanghai Pilot Program for Basic Research (No. 22TQ1400100-20).

\bibliography{custom}

\appendix
% \section{Appendix}
% \label{sec:appendix}
\clearpage

\section{Models}
\begin{itemize}
    \item \textbf{Qwen2.5-Instruct series.} We choose the \{0.5, 1.5, 3, 7, 14, 32, 72\} sizes.
    \item \textbf{Internlm2.5-Chat series.} We choose the \{1.8, 7, 20\} sizes.
    \item \textbf{Llama3.1-Instruct series.} We choose the \{8, 70\} sizes.
    \item \textbf{MiniCPM3-4B}~\footnote{\href{https://huggingface.co/openbmb/MiniCPM3-4B}{https://huggingface.co/openbmb/MiniCPM3-4B}}. It's a lightweight Chinese LLM.
    \item \textbf{Apollo series} We choose the Apollo2 \{0.5, 1.5, 7\} and Apollo {72}. Apollo models trained on Qwen and Qwen2 with a high quality medical dataset.
    \item \textbf{HuatuoGPT2 series} We choose the HuatuoGPT2 \{7, 14\}. We strictly followed the default load method of HuatuoGPT2-34B, but the inference time was too long, and the final results were not satisfactory. As a result, we did not conduct further experiments on HuatuoGPT2-34B.
\end{itemize}

\section{Training Details}
\label{appendix:training}
As shown in Figure~\ref{fig:combined} and Table~\ref{Time}, Qwen2.5-7B-Instruct is an excellent foundation model, considering both performance and time cost. However, for practical purposes, we prefer using a more lightweight model. We trained the model with LLaMA-Factory.~\footnote{\href{https://github.com/hiyouga/LLaMA-Factory}{https://github.com/hiyouga/LLaMA-Factory}} We use the default ds\_z3\_config and Lora fine-tuning. Detail parameters: per\_device\_train\_batch\_size: 3; gradient\_accumulation\_steps: 8; learning\_rate: 1.0e-5;num\_train\_epochs:4; lr\_scheduler\_type: cosine; warmup\_ratio: 0.1; fp16: true; ddp\_timeout: 180000000.

\section{Discussion on Test-Time Scaling}
% As shown in Table~\ref{Time}, we report the inference time of Qwen, which indicates that the CF-IR method can reduce the inference time in a zero-shot setting, while providing a demonstration that may increase the cost in a one-shot setting. Thus, a possible approach for practical application is leveraging strong LLM like GPT-4o to enhance the rule and utilizing the lightweight LLM during inference.

While test-time scaling like (CoT-SC, ToT or O1, R1) has attracted significant research attention, practical deployment in clinical settings requires careful consideration of GPU resource constraints, as hospitals typically have limited computational capacity. To address this, we quantitatively compare the inference efficiency of our method against CoT in Table~\ref{Time}, with measurements reported in GPU-hours.

\begin{table}
\centering  
\begin{tabular}{c|cc|cc}
\toprule
 & \multicolumn{2}{c|}{zero-shot}& \multicolumn{2}{c}{one-shot}\\ \hline
               & CoT& CF-IR& CoT&CF-IR\\ \hline
Qwen2.5-7B& 0.59& 0.46$\textcolor{blue}{\downarrow}$& 0.56&0.69$\textcolor{red}{\uparrow}$\\  
Qwen2.5-14B& 1.30& 1.26$\textcolor{blue}{\downarrow}$& 1.40&1.70$\textcolor{red}{\uparrow}$\\  
Qwen2.5-32B& 3.88& 3.60$\textcolor{blue}{\downarrow}$& 3.32&4.16$\textcolor{red}{\uparrow}$\\ 
 Qwen2.5-72B& 7.60&7.04$\textcolor{blue}{\downarrow}$& 5.80&6.84$\textcolor{red}{\uparrow}$\\ \bottomrule
 \end{tabular}

 \caption{Total Inference time of Qwen on CMQCIC-Bench across different methods. The unit is an hour$\cdot$GPU. The arrow indicates the change in inference time cost from CoT to CF-IR. }
    \label{Time}
\end{table}

\begin{table}
    \centering  
      % 或者使用 \scriptsize
    % \small
    \begin{tabular}{c|c c }
    \toprule
  Models& CoT& CF-IR
\\ \hline DeepSeek-V3&  3h40min&  3h29min$\textcolor{blue}{\downarrow}$\\ DeepSeek-R1& 10h10min& 9h30min$\textcolor{blue}{\downarrow}$\\ \bottomrule
 
    \end{tabular}
    \caption{Inference cost between DeepSeek-R1 and DeepSeek-V3. Experiments in zero-shot setting.}
    \label{deepseek time}
\end{table}

Comparing inference times between chat and reasoning models in Table~\ref{deepseek time}, we observe that while the reasoning model demonstrates better scalability, the performance gains remain marginal in Table~\ref{test-time scaling}. In the zero-shot setting, R1 did not demonstrate significant improvement. In contrast, our CF-IR framework outperformed the long cot method.

\begin{table}
    \centering  
      % 或者使用 \scriptsize
    % \small
    \begin{tabular}{c|c c c}
    \toprule
  & Standard&CoT&CF-IR\\ \hline Qwen2.5-7B&  66.49&  82.80& 82.92
\\ Qwen2.5-14B& 78.79& 86.49& 87.21
\\ 
Qwen2.5-32B&    75.54&    82.03& 86.11
\\  
 Qwen2.5-72B& 87.77& 87.51& 92.73
\\
 llama3.1-70B& 82.54& 85.85& 85.47
\\ \hline
 GPT-4o& 77.45& 88.91&91.84
\\
 DeepSeek-V3& 82.67& 86.83&91.84
\\
 DeepSeek-R1& 81.01& 82.99&92.73\\ \bottomrule
 
    \end{tabular}
    \caption{Performance on test-time scaling.}
    \label{test-time scaling}
\end{table}

% \begin{figure}[t]
% \centering
% \includegraphics[width=\linewidth]{distribution_data.png}
% \caption{The distribution of data in the CMQCIC-Bench dataset.}
% \label{fig:distribution_data}
% \end{figure}

\begin{table*}
    \centering  
    \small  % 或者使用 \scriptsize
    \begin{tabular}{c| c }
    \toprule
   Fileds Name&Explanation\\ \midrule
            definition&  The definition of the indicator\\ 
              formula&    The calculation formula of the indicator \\ significance&  The medical impact of indicator \\ 
      other&   The relative knowledge or supplement\\ 
              instruction\_standard&   The standard prompt for MQCIC with the rule  \\ 
     numerator&   The numerator of indicator \\ 
 denominator& The denominator of indicator \\ 
 rule&  The numerator rule of indicator \\ 
 facts  &The templated clinical facts list \\ 
 logical\_rules& The logical rules lists. Containing natural and symbolic languages.\\ \bottomrule
    \end{tabular}
    \caption{Main fields explanation of indicator file.}
    \label{indicator_File_fields}
\end{table*}

\begin{table*}
    \centering  
    \small  % 或者使用 \scriptsize
    \begin{tabular}{c |c }
    \toprule
   Fileds Name&Explanation\\ \midrule
            unique\_id&  The unique id of the indicator\\ 
              patient note&    The patient note of the instance\\ explaination&  The explanation of the answer\\ 
      label&   The label of the answer\\ 
              facts&   The list that contains all templated clinical facts with the related original text and answer\\ 
 logic& The list that contains logical rules with the answer\\ \bottomrule
    \end{tabular}
    \caption{Main fields explanation of data file.}
    \label{data_File_fields}
\end{table*}

% \subsubsection{Metrics}
\section{Prompt}
Here are the zero-shot prompt templates for the data construction, rule representation enhancement, clinical fact-based inferential rule reasoning method, and other prompt-based methods we used in this paper.

\begin{figure}[t]
\centering
\includegraphics[width=\linewidth]{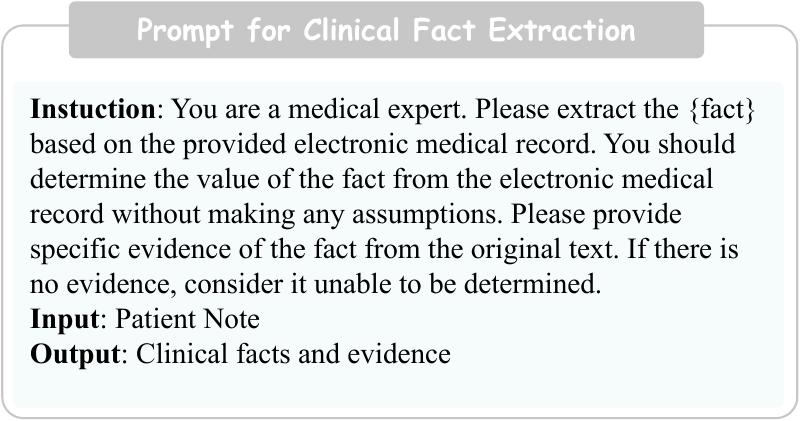}
\caption{The clinical fact extraction prompt.}
\label{fig:Prompt_CFE}
\end{figure}

For each indicator rule question, we provide a list of facts that should be extracted. The prompt for clinical fact extraction is shown in Figure~\ref{fig:Prompt_CFE}. The main fields are in the Table~\ref{indicator_File_fields} and Table~\ref{data_File_fields}.

\begin{figure}[t]
\centering
\includegraphics[width=\linewidth]{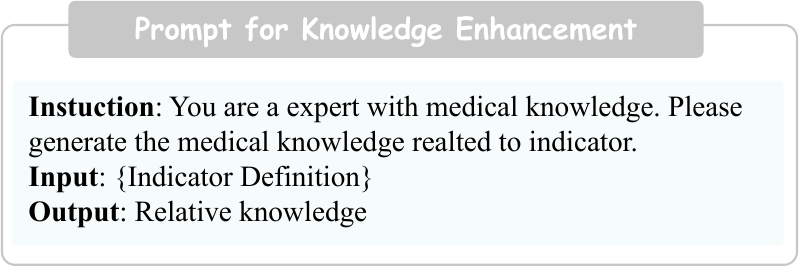}
\caption{The Prompt Template of Knowledge Enhancement.}
\label{fig:KE}
\end{figure}

\begin{figure}[t]
\centering
\includegraphics[width=\linewidth]{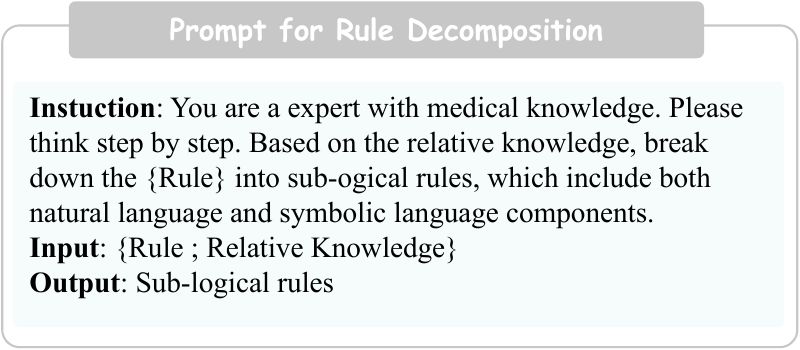}
\caption{The Prompt Template of Rule Decomposition.}
\label{fig:RD}
\end{figure}

\begin{figure}[t]
\centering
\includegraphics[width=\linewidth]{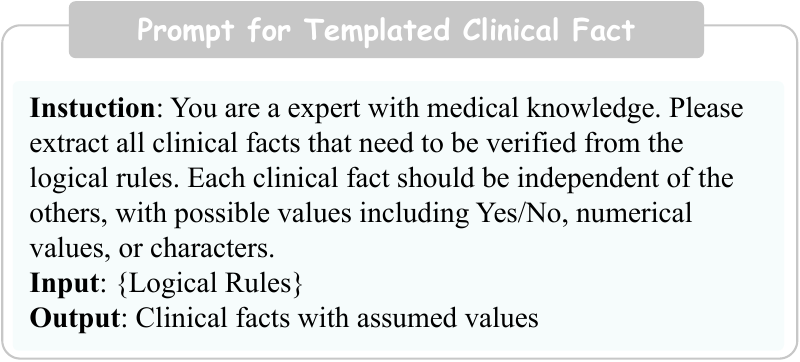}
\caption{The Prompt of Templated Clinical Fact.}
\label{fig:TCF}
\end{figure}

\begin{figure*}[t]
\centering
\includegraphics[width=\linewidth]{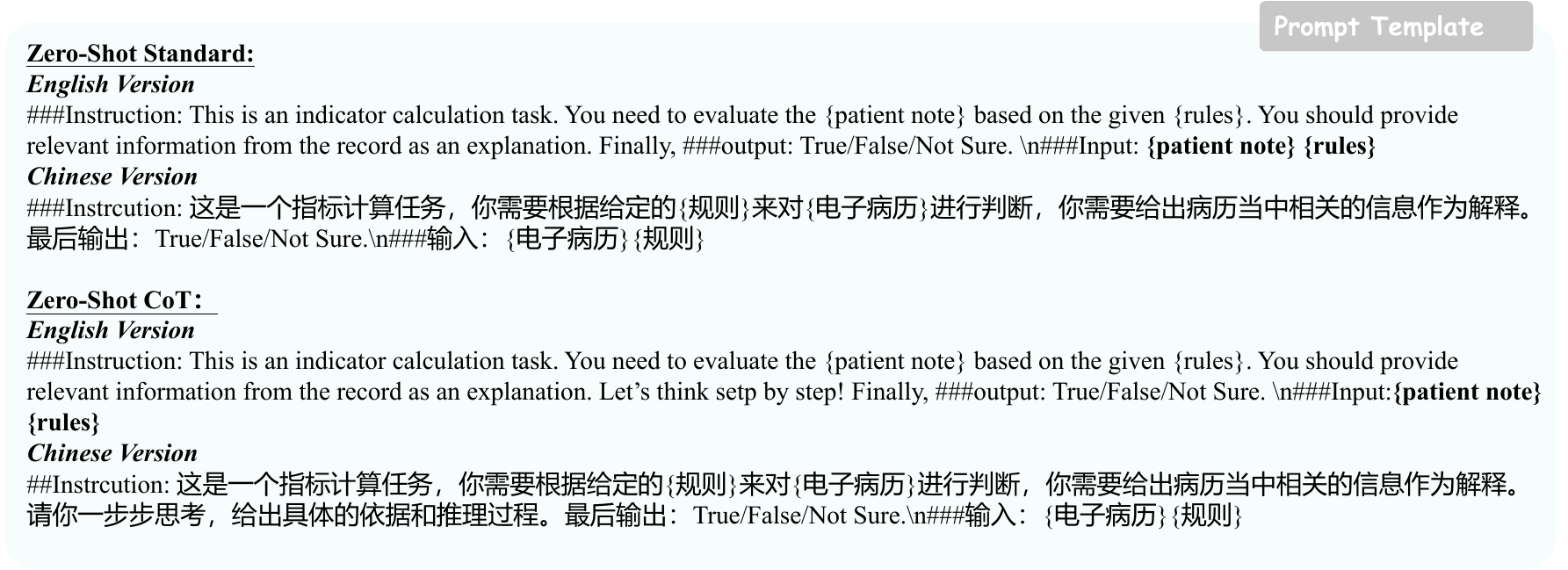}
\caption{The prompt template of standard and cot methods in translated English and Chinese version.}
\label{fig:Prompt_navie_cot}
\end{figure*}

\begin{figure*}[t]
\centering
\includegraphics[width=\linewidth]{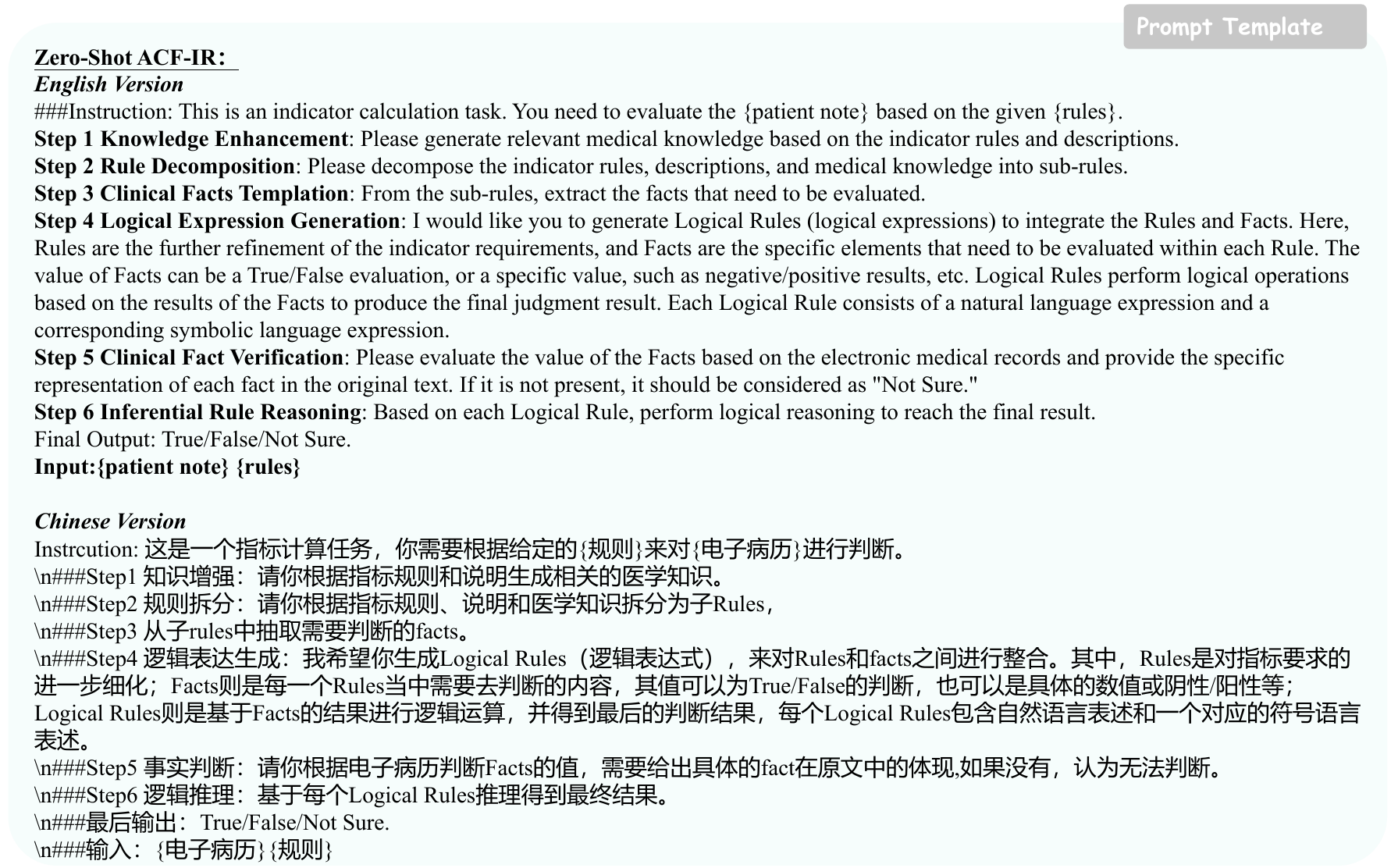}
\caption{The prompt template of ACF-IR method in translated English version and Chinese version.}
\label{fig:Prompt_ACF}
\end{figure*}

\begin{figure*}[t]
\centering
\includegraphics[width=\linewidth]{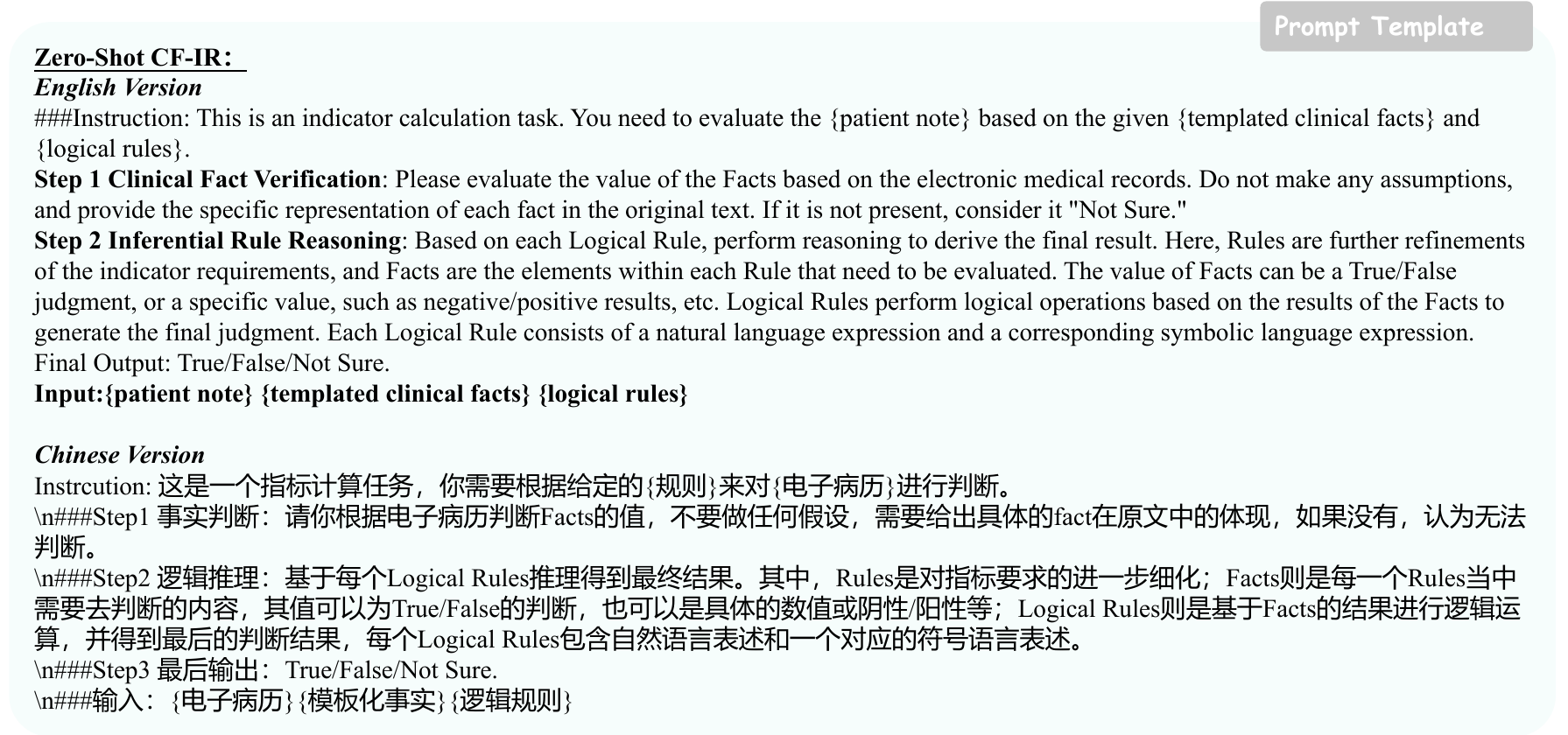}
\caption{The prompt template of CF-IR method in translated English and Chinese version.}
\label{fig:Prompt_CF}
\end{figure*}

\textbf{Prompt for Rule Enhancement}. As mentioned in Section 4, we leverage LLMs to transform rules through three steps: Knowledge Enhancement, Rule Decomposition, and Clinical Fact Templazation. As shown in Figure~\ref{fig:KE},~\ref{fig:RD} and~\ref{fig:TCF}, we leverage GPT-4o first to generate the relative knowledge, logical rules, and templated clinical facts separately. \textbf{Prompt for Different Methods}. The detail prompt of different method as shown in Figure~\ref{fig:Prompt_navie_cot}, Figure~\ref{fig:Prompt_ACF} and Figure~\ref{fig:Prompt_CF}.

% \newpage
\section{Additional Analysis Details}
\subsection{Error Analysis Details}
\label{apendix:error}
As shown in Table~\ref{cot_error} and Table~\ref{CF-IR_error} we observed: (1) Although the one-shot method reduces errors across both types in both approaches, the CoT method still results in more Type B errors, which may be due to the differing reasoning paths in the examples. (2) The CF-IR method effectively reduces Type B errors, but when it comes to Type A errors, the issue seems to be more related to the model's intrinsic capabilities, which our method has not been able to enhance or activate effectively.

\subsection{Clinical Fact Verification and Inferential Rule}
\label{appendix:FV} 
It may be due to the excessively strict scoring criteria that the overall score for clinical fact verification ability is relatively low, but the trend still aligns with expectations. As mentioned in Section~\ref{sec:ability analysis}, there is a clear correlation between the model parameters and capabilities in the Apollo and Qwen series. As shown in Table~\ref{ablation}, however, models like llama3.1 and HuatuoGPT2, due to differences in the number of parameters, fail to demonstrate this relationship.

% The faithfulness score difference compared to human experts is within 10\%, and the correctness score difference is within 5\%. Based on these results, we applied this method to evaluate the responses of other models.
 
\begin{table*}
    \centering  
    \small  % 或者使用 \scriptsize

  \resizebox{\textwidth}{!}{
    \begin{tabular}{c l|c c cc|cccc}
    \toprule
 & & \multicolumn{4}{c|}{Zero-shot CoT}&\multicolumn{4}{c}{One-shot CoT}\\ \hline  
         & &  Type A&  Type B& Type C& Total&Type A&Type B& Type C& Total\\ \hline  
\multirow{14}{*}{\rotatebox{90}{General}}&
MiniCPM3-4B
&    0.21 &    0.08 & 0.00 &  0.28 &0.13$\textcolor{blue}{\downarrow}$&   0.03$\textcolor{blue}{\downarrow}$& 0.00-&  0.16 
\\ \cline{2-10}
 & Internlm2.5-1.8B& 0.35 & 0.08 & 0.00 & 0.44 &0.23$\textcolor{blue}{\downarrow}$& 0.08-& 0.00-& 0.31 
\\
 & Internlm2.5-7B& 0.15 & 0.12 & 0.00 & 0.27 &0.13$\textcolor{blue}{\downarrow}$& 0.03$\textcolor{blue}{\downarrow}$& 0.00-& 0.16 
\\
 & Internlm2.5-20B& 0.17 & 0.05 & 0.00 & 0.22 &0.11$\textcolor{blue}{\downarrow}$& 0.03$\textcolor{blue}{\downarrow}$& 0.00-& 0.14 
\\   \cline{2-10}
 &Qwen2.5-0.5B&    0.34 &    0.10 & 0.00 & 0.44 &0.29$\textcolor{blue}{\downarrow}$&   0.10-& 0.00-& 0.39 
\\   
 &Qwen2.5-1.5B&    0.21 &    0.16 & 0.00 & 0.37 &0.24$\textcolor{red}{\uparrow}$&   0.05$\textcolor{blue}{\downarrow}$& 0.00-& 0.29 
\\  
  &Qwen2.5-3B&    0.16 &    0.11 & 0.00 & 0.27 &0.15$\textcolor{blue}{\downarrow}$&   0.07$\textcolor{blue}{\downarrow}$& 0.00-& 0.22 
\\   
         &Qwen2.5-7B&     0.13 &     0.04 & 0.00 & 0.17 &0.12$\textcolor{blue}{\downarrow}$&   0.03$\textcolor{blue}{\downarrow}$& 0.00-& 0.15 
\\   
         &Qwen2.5-14B&     0.10 &     0.08 & 0.00 & 0.18 &0.08$\textcolor{blue}{\downarrow}$&   0.04$\textcolor{blue}{\downarrow}$& 0.00-& 0.12 
\\
 & Qwen2.5-32B&    0.09 &    0.05 & 0.00 & 0.14 &0.06$\textcolor{blue}{\downarrow}$&    0.04$\textcolor{blue}{\downarrow}$& 0.00-& 0.10 
\\   
         &Qwen2.5-72B&     0.09 &     0.03 & 0.00 & 0.12 &0.07$\textcolor{blue}{\downarrow}$&   0.02$\textcolor{blue}{\downarrow}$& 0.00-& 0.09 
\\   \cline{2-10}
         &llama3.1-8B
&     0.23 &     0.14 & 0.00 & 0.37 &0.14$\textcolor{blue}{\downarrow}$&   0.04$\textcolor{blue}{\downarrow}$& 0.00-& 0.18 
\\
&llama3.1-70B
&     0.10 &     0.04 & 0.00 & 0.14 &0.08$\textcolor{blue}{\downarrow}$&   0.03$\textcolor{blue}{\downarrow}$& 0.00-& 0.11 
\\  \cline{2-10}
&GPT-4o&     0.08 &     0.03 & 0.00 & 0.11 &0.06$\textcolor{blue}{\downarrow}$&   0.02$\textcolor{blue}{\downarrow}$& 0.00-& 0.08 
\\ 
         \hline  
         \multirow{6}{*}{\rotatebox{90}{Medical}}  
         &HuatuoGPT2-7B
&     0.17 &     0.28 & 0.00 & 0.46 &0.23$\textcolor{red}{\uparrow}$&   0.23$\textcolor{blue}{\downarrow}$& 0.02$\textcolor{red}{\uparrow}$& 0.47 
\\
         &HuatuoGPT2-14B&     0.32 &     0.13 & 0.01 & 0.45 &0.28$\textcolor{blue}{\downarrow}$&   0.18$\textcolor{red}{\uparrow}$& 0.01-& 0.48 
\\ \cline{2-10}
 & Apollo2-0.5B&    0.34 &    0.19 & 0.04 & 0.58 &0.30$\textcolor{blue}{\downarrow}$&   0.15$\textcolor{blue}{\downarrow}$& 0.02$\textcolor{blue}{\downarrow}$& 0.46 
\\
 & Apollo2-1.5B&    0.23 &    0.25 & 0.00 & 0.48 &0.27$\textcolor{red}{\uparrow}$&   0.07$\textcolor{blue}{\downarrow}$& 0.00-& 0.34 
\\
         &Apollo2-7B&     0.17 &     0.23 & 0.00 & 0.40 &0.24$\textcolor{red}{\uparrow}$&   0.04$\textcolor{blue}{\downarrow}$& 0.00-& 0.28 
\\ 
 &Apollo-72B&    0.14 &    0.10 & 0.00 & 0.24 &0.11$\textcolor{blue}{\downarrow}$&   
   0.03$\textcolor{blue}{\downarrow}$& 0.00-& 0.14 
\\ \hline  
 
    \end{tabular}
    }
    \caption{Error type distribution of LLMs on CMQCIC-Bench dataset. Arrows represent the changes from zero-shot to one-shot.}
    \label{cot_error}
\end{table*}
\begin{table*}
    \centering  
    \small  % 或者使用 \scriptsize

  \resizebox{\textwidth}{!}{
    \begin{tabular}{c l|c c cc|cccc}
    \toprule
 & & \multicolumn{4}{c|}{Zero-shot CF-IR}&\multicolumn{4}{c}{One-shot CF-IR}\\ \hline  
         & &  Type A&  Type B& Type C& Total&Type A&Type B& Type C& Total\\ \hline  
\multirow{14}{*}{\rotatebox{90}{General}}&
MiniCPM3-4B
&    0.27 
&    0.04 
& 0.00 
&  0.31 
&0.11$\textcolor{blue}{\downarrow}$&   0.06$\textcolor{red}{\uparrow}$& 0.00-&  0.17 
\\ \cline{2-10}
 & Internlm2.5-1.8B& 0.39 
& 0.07 
& 0.00 
& 0.46 
&0.27$\textcolor{blue}{\downarrow}$& 0.10$\textcolor{red}{\uparrow}$& 0.00-& 0.36 
\\
 & Internlm2.5-7B& 0.18 
& 0.03 
& 0.00 
& 0.21 
&0.13$\textcolor{blue}{\downarrow}$& 0.02$\textcolor{blue}{\downarrow}$& 0.00-& 0.16 
\\
 & Internlm2.5-20B& 0.15 
& 0.04 
& 0.00 
& 0.19 
&0.10$\textcolor{blue}{\downarrow}$& 0.01$\textcolor{blue}{\downarrow}$& 0.00-& 0.11 
\\   \cline{2-10}
 &Qwen2.5-0.5B&    0.28 
&    0.18 
& 0.00 
& 0.46 
&0.28-&   0.18-& 0.01$\textcolor{red}{\uparrow}$& 0.46 
\\   
 &Qwen2.5-1.5B&    0.36 
&    0.02 
& 0.00 
& 0.38 
&0.25$\textcolor{blue}{\downarrow}$&   0.02-& 0.00-& 0.27 
\\  
  &Qwen2.5-3B&    0.31 
&    0.01 
& 0.00 
& 0.32 
&0.16$\textcolor{blue}{\downarrow}$&   0.02$\textcolor{red}{\uparrow}$& 0.00-& 0.18 
\\   
         &Qwen2.5-7B&     0.14 
&     0.03 
& 0.00 
& 0.17 
&0.08$\textcolor{blue}{\downarrow}$&   0.01$\textcolor{blue}{\downarrow}$& 0.00-& 0.10 
\\   
         &Qwen2.5-14B&     0.12 
&     0.01 
& 0.00 
& 0.13 
&0.07$\textcolor{blue}{\downarrow}$&   0.01-& 0.00-& 0.08 
\\
 & Qwen2.5-32B&    0.12 
&    0.02 
& 0.00 
& 0.14 
&0.05$\textcolor{blue}{\downarrow}$&    0.00$\textcolor{blue}{\downarrow}$& 0.00-& 0.05 
\\   
         &Qwen2.5-72B&     0.06 
&     0.01 
& 0.00 
& 0.07 
&0.03$\textcolor{blue}{\downarrow}$&   0.01-& 0.00-& 0.04 
\\   \cline{2-10}
         &llama3.1-8B
&     0.20 
&     0.02 
& 0.00 
& 0.22 
&0.12$\textcolor{blue}{\downarrow}$&   0.01$\textcolor{blue}{\downarrow}$& 0.00-& 0.14 
\\
&llama3.1-70B
&     0.12 
&     0.03 
& 0.00 
& 0.15 
&0.07$\textcolor{blue}{\downarrow}$&   0.00$\textcolor{blue}{\downarrow}$& 0.00-& 0.08 
\\  \cline{2-10}
&GPT-4o&     0.07 
&     0.01 
& 0.01 
& 0.08 
&0.05$\textcolor{blue}{\downarrow}$&   0.01-& 0.00$\textcolor{red}{\uparrow}$& 0.06 
\\ 
         \hline  
         \multirow{6}{*}{\rotatebox{90}{Medical}}  
         &HuatuoGPT2-7B
&     0.33 
&     0.16 
& 0.01 
& 0.50 
&0.31$\textcolor{blue}{\downarrow}$&   0.11$\textcolor{blue}{\downarrow}$& 0.02$\textcolor{red}{\uparrow}$& 0.43 
\\
         &HuatuoGPT2-14B&     0.43 
&     0.09 
& 0.02 
& 0.54 
&0.46$\textcolor{red}{\uparrow}$&   0.08$\textcolor{blue}{\downarrow}$& 0.04$\textcolor{red}{\uparrow}$& 0.57 
\\ \cline{2-10}
 & Apollo2-0.5B&    0.32 
&    0.15 
& 0.12 
& 0.59 
&0.27$\textcolor{blue}{\downarrow}$&   0.07$\textcolor{blue}{\downarrow}$& 0.01$\textcolor{blue}{\downarrow}$& 0.35 
\\
 & Apollo2-1.5B&    0.30&    0.17& 0.01 
& 0.49 
&0.26$\textcolor{blue}{\downarrow}$&   0.08$\textcolor{blue}{\downarrow}$& 0.01-& 0.35 
\\
         &Apollo2-7B&     0.29 
&     0.10 
& 0.00 
& 0.38 
&0.14$\textcolor{blue}{\downarrow}$&   0.21$\textcolor{red}{\uparrow}$& 0.01-& 0.35 
\\ 
 &Apollo-72B&    0.26 
&    0.01 
& 0.00 
& 0.27 
&0.11$\textcolor{blue}{\downarrow}$&   
   0.03$\textcolor{red}{\uparrow}$& 0.00-& 0.14 
\\ \hline  
 
    \end{tabular}
    }
    \caption{Error type distribution of LLMs on CMQCIC-Bench dataset. Arrows represent the changes from zero-shot to one-shot.}
    \label{CF-IR_error}
\end{table*}

\begin{table*}
    \centering  
  \resizebox{\textwidth}{!}{
    \begin{tabular}{c l|c c |c c|c}
    \toprule
 & & \multicolumn{2}{c|}{Clinical Fact Verification}& \multicolumn{2}{c|}{Inferential Rule Reasoning}&One-Stage\\ \hline  
         & &  Faithfulness&  Correctness
& NL(ACC)&SY(ACC)& Best(ACC)\\ \hline  
 % \multicolumn{6}{c}{General Models}\\ \hline  
\multirow{10}{*}{\rotatebox{90}{General}}&
MiniCPM3-4B
& 45.14& 36.05
& 67.38 &69.68 
& 72.10*
\\   
 &Qwen2.5-0.5B& 45.37& 40.1
& 74.90 &75.92 
& 56.05*
\\   
 &Qwen2.5-1.5B& 46.45& 43.67
& 75.66 &74.64 
& 66.11*
\\  
  &Qwen2.5-3B& 52.08& 50.02
& 83.69 &81.18 
& 76.05*
\\   
         &Qwen2.5-7B&  67.63&  53.13
&  90.44 &88.91 
& 82.92*
\\   
         &Qwen2.5-14B&  67.94&  62.35
&  92.35 &94.52 
& 87.21*
\\
 & Qwen2.5-32B& 75.41& 69.47
& 89.80 & 94.14 
&86.11*
\\   
         &Qwen2.5-72B&  74.07&  77.13
&  93.63 &93.88 
& 92.73*
\\   
         &llama3.1-8B
&  47.89&  37.88
&  82.67 &80.89 
& 78.34*
\\
&llama3.1-70B
&  48.59&  40.48
&  85.60 &85.98 
& 85.47*
\\ 
         \hline  
         \multirow{6}{*}{\rotatebox{90}{Medical}}  
         &HuatuoGPT2-7B
&  13.27&  22.93
&  47.89 &44.20 
& 54.26*
\\
         &HuatuoGPT2-14B&  32.55&  35.47
&  62.42 &62.03 
& 55.28*
\\
 & Apollo2-0.5B& 5.19& 10.73
& 9.80 &18.47 
& 49.29*
\\
 & Apollo2-1.5B& 25.31& 29.21
& 48.53 &52.86 
& 55.03*
\\
         &Apollo2-7B&  33.78&  33.93
&  56.30 &42.29 
& 61.91*
\\ 
 &Apollo-72B& 42.31& 32.07
& 
76.81 &   
71.59 
& 
76.24*
\\ \hline  
 \multicolumn{2}{c|}{Average}& 45.18& 42.16& \textbf{71.12}& 70.69
& 
70.94*\\ \hline
    \end{tabular}
    }
    \caption{Performance of Clinical Fact Verification and Inferential Rule Reasoning on CMQCIC-Bench. For Clinical Fact Verification, we utilize DeepSeek to assess both faithfulness and correctness. NL represents Natural Language, SY denotes Symbolic Language, and ACC stands for Accuracy. To enable a clearer comparison, we present the best results of the "standard," "CoT," and "CF-IR" approaches in Table~\ref{main_result}. All experiments were conducted in the zero-shot setting.}
    \label{ablation}
\end{table*}

% \subsection{Case Study on CMQCIC-Bench Private}
% As shown in Figure

% \begin{figure*}[t]
% \centering
% \includegraphics[width=\linewidth]{f}
% \caption{The prompt template of CF-IR method in English and Chinese version.}
% \label{fig:}
% \end{figure*}

\end{document}